# Examining Linguistic Shifts in Academic Writing Before and After the Launch of ChatGPT: A Study on Preprint Papers

Tong Bao[1], Yi Zhao[1], Jin Mao[2], Chengzhi Zhang[1, *]

1. Department of Information Management, Nanjing University of Science and Technology, Nanjing, 210094, China
2. School of Information Management, Wuhan University, Wuhan 430072, China

**Abstract:** Large Language Models (LLMs), such as ChatGPT, have prompted academic concerns about their impact on academic writing. Existing studies have primarily examined LLM usage in academic writing through quantitative approaches, such as word frequency statistics and probability-based analyses. However, few have systematically examined the potential impact of LLMs on the linguistic characteristics of academic writing. To address this gap, we conducted a large-scale analysis across 823,798 abstracts published in last decade from arXiv dataset. Through the linguistic analysis of features such as the frequency of LLM-preferred words, lexical complexity, syntactic complexity, cohesion, readability and sentiment, the results indicate a significant increase in the proportion of LLM-preferred words in abstracts, revealing the widespread influence of LLMs on academic writing. Additionally, we observed an increase in lexical complexity and sentiment in the abstracts, but a decrease in syntactic complexity, suggesting that LLMs introduce more new vocabulary and simplify sentence structure. However, the significant decrease in cohesion and readability indicates that abstracts have fewer connecting words and are becoming more difficult to read. Moreover, our analysis reveals that scholars with weaker English proficiency were more likely to use the LLMs for academic writing, and focused on improving the overall logic and fluency of the abstracts. Finally, at discipline level, we found that scholars in Computer Science showed more pronounced changes in writing style, while the changes in Mathematics were minimal.

---

* Corresponding author: Chengzhi Zhang (zhangcz@njust.edu.cn).

## 1 Introduction

ChatGPT, a chatbot launched by OpenAI in November 2022, is considered one of the most powerful and widely adopted Large Languages Models (LLMs) to date. ChatGPT can generate high-quality text based on user-provided context or prompts, and its free accessibility has led to significant impacts in both industrial and academic (Cao et al., 2023). In scientific research, several studies have shown that ChatGPT have been widely used in various aspects of academic writing, including drafting manuscripts, summarizing abstracts, and generating reviews (Geng & Trotta, 2024; Huang & Tan, 2023). However, the increasing reliance on LLMs in academic writing has raised concerns regarding content originality, authorship attribution, and the potential transformation of academic writing styles (Thorp, 2023). For instance, in top-tier computer science conferences, 6.5% to 16.9% of peer review comments may have undergone significant modifications by LLMs (Liang, et al., 2024a). Since LLM-generated text is difficult to trace back to specific sources, it increases the risk of unintentional plagiarism and the dissemination of biased information, challenging the objectivity of academic writing (AlAfnan & MohdZuki, 2023). Furthermore, from a broader academic culture perspective, academic disciplines have historically developed unique writing conventions that shape the communication of knowledge within their respective fields (Song et al., 2023). However, the widespread use of LLMs may blur these disciplinary boundaries, leading to a loss of stylistic distinctiveness in academic writing.

In response to these concerns, academic institutions and journals have introduced policies to regulate LLM usage in scholarly publications. For example, renowned journals ***Science*** and ***Nature*** have updated their submission guidelines, stating that they do not recognize LLMs as contributing authors and require researchers to disclose any use of AI in the acknowledgments section (Thorp, 2023). Additionally, recent studies have employed AI-detection tools or word frequency statistics to investigate LLM usage in academic writing, revealing that LLM-generated content often increases the prevalence and frequency of certain word choices (Hwang et al., 2024; Juzek & Ward, 2025; Liang et al., 2024a; Liu & Bu, 2024). While these studies provide valuable insights, they primarily focus on identifying AI-generated text rather than examining how LLMs influence the linguistic characteristics of academic writing. Academic papers serve not only as a medium for publishing the latest research findings but also a semantic representation of knowledge, while linguistic features describe the lexical and grammatical choices of scholars. Specifically, linguistic features, such as vocabulary choice, sentence structure, and writing patterns, play a critical role in distinguishing human-authored work from AI-generated text (Liang et al., 2024a). Thus, analyzing changes in linguistic features of academic writing offers not only a new perspective to assess the authenticity and originality of academic content, but also helps explore the impact of LLMs on the overall writing style of the academic community.

To the best of our knowledge, no existing research has systematically analyzed how academic writing styles are evolving in the era of LLMs from a linguistic perspective. Therefore, to fill this gap, we take ChatGPT, one of the most widely used LLMs, as an example to investigate how the linguistic features of academic papers are evolving. This study provides empirical evidence of LLMs' influence on academic writing and offers valuable insights for the academic community, particularly in the development of AI detection tools and the formulation of policies for responsible LLM usage. To achieve this goal, we aim to answer the following research question:

***RQ1***: Has the linguistic style of academic writing changed before and after the ChatGPT era? If so, what changes have occurred?

With the growing trend of international collaboration in academic research, non-native English-speaking scholars (NNESs) are playing an increasingly important role in academic publications (Gök & Karaulova, 2024; Gui et al., 2019; Ribeiro et al., 2018). However, NNESs often face challenges in academic publication due to language barriers (Flowerdew, 1999; Huang, 2010). For example, NNESs tend to exhibit a more limited vocabulary, lower word accuracy, and less fluent sentence construction, whereas native English-speaking scholars (NESs) are more adept at writing longer, more coherent sentences and well-structured paragraphs (Ortega, 2003). To promote equity in academic publications, mainstream academic publishers have taken proactive steps to provide professional language editing services for NNESs. However, there are still a large proportion of scholars who may not be able to afford the cost of language editing services due to financial constraints. Recently, LLMs like ChatGPT have emerged as an cost-effective solution to help NNESs overcome language barriers in academic writing (Cao et al., 2023). These models provide real-time assistance in word choice, grammar correction, and language polishing, enabling scholars to express their viewpoints more fluently and accurately in academic writing (Lozić & Štular, 2023). However, whether the use of LLMs introduces notable differences in writing style changes between NESs and NNESs remains an open question. Therefore, our second research question is as follows:

***RQ2***: Are there differences in writing style changes between NNESs and NESs? If so, what are the differences?

At present, LLMs have been widely applied and achieved success in the field of computer science, but their application in other disciplines may still be underdeveloped (Onal & Kulavuz-Onal, 2024). For example, in fields such as agriculture, finance, and the arts, LLMs may require more customization and domain-specific knowledge to achieve optimal results. Similarly, in scientific research, differences in academic backgrounds may lead to varying levels of acceptance and proficiency in using LLMs (Liang, et al., 2024b). Given these variations, it is crucial to explore whether writing style changes induced by LLMs differ across disciplines. Therefore, we further propose our third research question:

***RQ3***: What are the differences in writing style variations between disciplines?

## 2 Literature review

### *2.1 Writing style of academic papers*

Academic papers serve as the primary medium for expression and dissemination of scientific knowledge, playing a crucial role in scientific research and development. Unlike news articles or novels, academic writing is distinguished by its focus on clear communication of knowledge, aiming to present research results and findings with accuracy and precision. At the same time, while academic writing is often characterized by structural and phrasal complexity, concise and precise language, simple sentence structure and fluency of expression are widely recognized standards of academic writing style (Biber & Gray, 2010).

The writing style of academic papers is shaped by various factors, including a scholar's language background, writing and training experience, and the paper's type or target audience (Atkinson, 1998; Gross et al., 2002). Recently, the development of LLMs, particularly generative pre-trained language models like ChatGPT and LLaMa, has had a significant impact on academic writing. These models have notably improved the performance of machine translation and text generation capabilities, which contributes to the accessibility and efficiency of academic writing, especially for scholars with limited English writing skills (Liu & Bu, 2024).

Although the writing style is not the primary factor directly determining the innovativeness of academic paper, it plays a crucial role in its dissemination and impact (Chen et al., 2020; Lu et al., 2019). An appropriate writing style enhance a paper's readability and appeal, potentially increasing its downloads and citations (Hu et al., 2021). Moreover, writing style influences how readers comprehend and evaluate the paper's innovation, subsequently impacting its academic impact and recognition. In essence, a well-crafted writing style helps readers better understand the research more easily and thus better evaluate the innovation and contribution of the paper, which is beneficial for enhancing the academic reputation and influence of authors.

### *2.2 Writing style measures of academic papers*

Academic papers exhibit distinct linguistic characteristics that set them apart from other text forms (Bennett, 2009). Previous studies have analyzed the writing style of academic papers across multiple dimensions, primarily focusing on word-level, sentence-level, and paragraph-level features (Dong et al., 2024; Song et al., 2023). At the word level, measures such as lexical density and lexical sophistication assess different aspects of lexical complexity in academic papers. While lexical density evaluates the proportion of content words in a text (Greenhalgh et al., 2023), lexical sophistication reflects the depth and richness of the vocabulary used. At the sentence level, syntactic complexity captures the variety and logic structure of expression, commonly measured through the number of clauses, T-units, and sentences (Wang et al., 2023). Furthermore, academic papers commonly follow the IMRaD structure (Introduction, Methods, Results, and Discussion), which helps readers understand the logical framework and section arrangement (Gidiotis & Tsoumakas, 2020). For instance, the introduction section presents relevant background and previous works, whereas the results section presents a study's findings which may include statistical results and graphical representations. Finally, the readability of a paper, which encompasses vocabulary usage and sentence structure, serves as a crucial indicator of both the presentation of the paper and the reader's comprehension (Plavén-Sigray et al., 2017; Snow, 2010).

As scholars continue to study the writing styles of academic papers, numerous tools have been developed to enhance the computational efficiency and comprehensiveness of linguistic metrics. Lu (2010) developed L2SCA, a widely used tool in corpus linguistics, which incorporates fourteen different measures derived from the second language development literature to assess syntactic complexity. Based on this work, Kyle & Crossley (2018) introduced new computational metrics and developed tools such as TAACO (Crossley et al., 2016) and TAACS (Kyle & Crossley, 2015) for calculating syntactic complexity and sentence cohesion. Additionally, the use of custom word lists combined with Python or R packages are also convenient methods of analysis (Lu et al. 2019). These research and tools have enriched the methodology and application of language stylistic studies, providing a foundation for further research on writing styles in the field of academic writing. In this study, we expanded on existing indicators by incorporating the scholars' sentiment scores and calculating the number of high-frequency LLMs-preferred words to further improve the measurement system of writing style in the era of LLMs.

### *2.3 Application of LLMs in academic writing*

With the advancement of deep learning techniques, various language editing tools such as Google Translate (Mundt & Groves, 2016) and Grammarly (Nur Fitria, 2021) have been developed to enhance the quality of academic writing, which target spelling errors, rhetorical expression, and sentence coherence. However, even if the revised text is correct in terms of vocabulary and grammar, it may still not meet the stylistic requirements of academic writing. Recent studies suggest that LLMs can be used for grammar correction, literature summarization, and the generation of abstracts and review papers, making them powerful tools for addressing the aforementioned challenges (Altmäe et al., 2023; Gao et al., 2023; Lozić & Štular, 2023).

The scientific community has evaluated the capabilities of LLMs in academic writing to explore their strengths and limitations (Ma et al., 2023). It is widely recognized that papers or summaries generated by LLMs generally exhibit satisfactory grammar, rhetorical quality, and logical coherence. However, the generated content often struggles with authenticity and originality (Uzun, 2023). As a result, ongoing discussions emphasize the need for greater transparency in academic writing, with suggestions that texts generated by LLMs should explicitly indicate their source to prevent potential ethical and academic integrity issues. For example, some academic journals require papers to clearly mark which parts are generated by LLMs to bolster the transparency and credibility of scientific research. Meanwhile, certain academic institutions are considering the establishment of ethical review mechanisms to rigorously assess papers involving LLM-generated content, ensuring they comply with academic ethics and standards (Thorp, 2023).

Several studies have quantitatively analyzed the impact of LLMs on academic writing. Geng & Trotta (2024) analyzed word frequency changes in abstracts and compared original abstracts with those modified by LLMs, highlighting the growing influence of LLMs in academic writing. Additionally, Liang et al. (2024a) introduced a distributed quantification framework to explore the relationship between LLM usage, research background, and author characteristics, revealing that authors in well-established fields and with higher publication frequencies were more likely to adopt LLMs. Furthermore, Liu & Bu (2024) employed AI detection tools to assess the likelihood of AI-generated content in preprint abstracts, observing an increase in this probability following the introduction of ChatGPT.

Despite these insights, prior studies have mainly focused on quantitative analyses and probability measures, with limited attention to textual linguistic features and their variations. In-depth linguistic analyses can provide a more detailed understanding of how LLMs influence word choice, structural changes, and the evolution of academic writing style. To the best of our knowledge, no comprehensive study has yet analyzed the impact of LLMs on academic writing from a linguistic perspective, and this study aims to fill this research gap.

# 3 Methodology

## *3.1 Data collection*

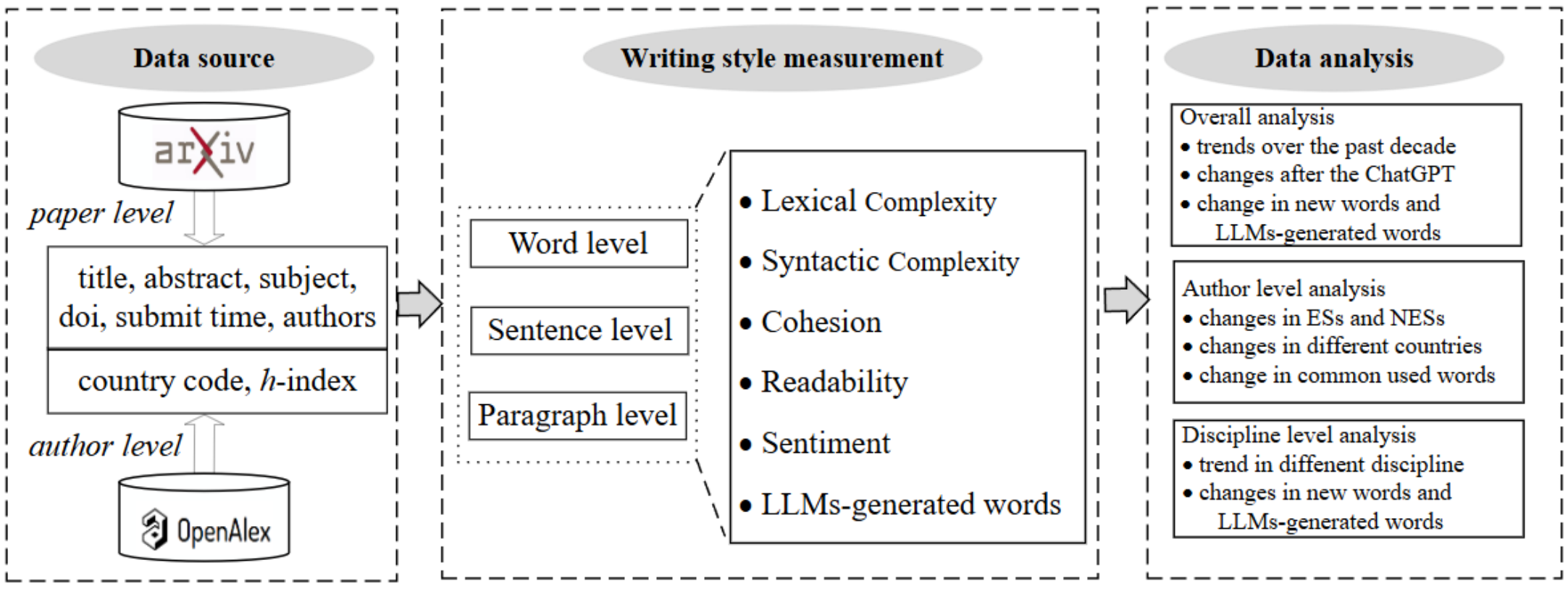

Fig. 1. Framework of this study

We collected metadata on a combined total of 1, 582, 837 articles published on arXiv[1] between 2014 and 2023 via the Kaggle platform[2], including information such as title, abstract, author list, subject category, digital object identifier (DOI), and submit time. The abstracts of all the articles were used as the primary data for the study and data with missing author lists, title, subject category, submission time or abstract length less than 50 were removed. ChatGPT was released on November 30, 2022, and given that the diffusion of emerging technologies and user adoption requires a transitional phase, there is uncertainty regarding the influence of ChatGPT on articles published in the short term. Therefore, following technology diffusion theory commonly used in social economics (Rogers et al., 2014; Greenhalgh et al., 2004), we excluded data from December 2022 to mitigate the impact of short-term fluctuations and ensure a more accurate representation of the evolution of writing styles over time. Finally, a total of 905, 057 abstracts were retained. To better display the changes, the data was divided into four time periods for each year, resulting in a total of 40 time periods. Of these, 36 time periods occurred prior to the release of ChatGPT, while the remaining four occurred after its release. Figure 1 presents the overall framework of this study.

(1) Discipline Classification

To investigate the variations in writing styles across different disciplines, each article was categorized according to the subject list provided in the arXiv metadata. For articles with multiple subject categories, the first-listed subject was designated as the primary category. In the end, we identified seven subjects including *Computer Science*, *Electrical Engineering and Systems Science*, *Mathematics*, *Physics*, *Quantitative Biology*, *Quantitative Finance and Statistics*. Among these, articles from *Computer Science*, *Physics*, and *Mathematics* accounted for 823798 records, representing over 91% of the total dataset (905,057 records). This substantial proportion underscores their significance within the dataset. Therefore, in this study, we removed abstracts outside of these three disciplines to ensure a more representative sample for our analysis. The data distribution is shown in Table 1.

TABLE. 1. Number of articles in different disciplines

| Year | Computer Science | Physics | Mathematics | Total | Ratio (%) |
|---|---|---|---|---|---|
| 2014 | 8,007 | 16,095 | 24,485 | 48,587 | 5.90 |
| 2015 | 9,932 | 17,473 | 27,028 | 54,433 | 6.61 |
| 2016 | 12,956 | 18,371 | 27,985 | 59,312 | 7.20 |
| 2017 | 17,251 | 19,709 | 30,176 | 67,136 | 8.15 |
| 2018 | 23,012 | 20,774 | 32,099 | 75,885 | 9.21 |
| 2019 | 28,270 | 21,950 | 33,467 | 83,687 | 10.16 |
| 2020 | 34,344 | 22,976 | 35,862 | 93,182 | 11.31 |
| 2021 | 38,879 | 22,280 | 36,146 | 97,305 | 11.81 |
| 2022 | 45,265 | 22,428 | 37,738 | 105,431 | 12.80 |
| 2023 | 61,742 | 24,826 | 52,272 | 138,840 | 16.85 |
| Total | 279,658 | 206,882 | 337,258 | 823,798 | 100 |

(2) Author-Level language Identification

In the scientific community, authorship order is often an indicator of relative contribution, with the first author typically making the most substantial contribution to the research and writing, while the corresponding author generally serves as a senior researcher or supervisor (Corrêa Jr et al., 2017; Larivière et al., 2016; Lu et al., 2022; Ueda et al., 2021). Given the absence of corresponding author information in the arXiv metadata and the strong correlation between corresponding authors and last authors (Fine & Kurdek, 2017; Dance, 2012), we randomly selected 2,000 articles and found that in 1,520 (76%) of them, the first author and the last author were from the same country. This percentage does not even include papers where the first author and last author were from the different countries but had the same linguistic background (NNESs or NESs). This suggests that, in most cases, the first and last authors share similar linguistic influences. Therefore, in this study, given the primary role of the first author in manuscript drafting, we considered the first author to be the most influential individual in shaping the academic writing style.

In this paper, we determine the linguistic background of the first author based on the country where they have published the most papers. Since institutional affiliations and countries may change over an academic career, we retrieved the historical institutional affiliation information of all authors in our dataset. The results showed that only 12% of authors had published papers linked to institutions in more than two countries. Because we define an author's linguistic background based on their country of affiliation, even if these 12% of authors had experienced cross-country mobility, for example, moving between different English-speaking countries, such changes would generally wouldn't affect linguistic background and thus still fall within the scope of our study.

To determine the country of the first author, we employed the Openalex API[3] to retrieve the author ID and country code for each article. In addition, in order to mitigate potential ambiguities from searching the paper's title directly, we used two strategies to enhance match accuracy. For articles with a DOI in the metadata, we directly retrieved the author ID and country code through a DOI search. For articles without a DOI, we matched the article title, author's name, and publication year to determine the necessary information. Articles that did not align using either strategy were excluded from the analysis. Authors were then categorized as NNESs or native NESs based on the common spoken language of their country (Melitz & Toubal, 2014), as shown in Table 2.

TABLE. 2. Country groupings for NESs and NNESs

| Group | Examples | Articles | Ratio (%) |
|---|---|---|---|
| NESs | UNITED KINGDOM, UNITED STATES, CANADA, AUSTRALIA, NEW ZEALAND | 273275 | 33.2% |
| NNESs | FRANCE, ITALY, NETHERLANDS, NORWAY, GERMANY, CHINA, INDIA, RUSSIA, JAPAN, KOREA | 550523 | 66.8% |

*3.2 Measurement of writing style in academic papers*

In this study, to comprehensively evaluate linguistic changes in academic writing style, we drew on the studies of Dong et al. (2024), Song et al. (2023), and Lu et al. (2019), who constructed evaluation frameworks from different perspectives to assess academic writing style. Building upon their work, we supplemented and designed a set of multidimensional metrics to assess stylistic shifts in arXiv abstracts. These frameworks cover multiple levels of textual analysis, including word-level, phrase-level, and sentence-level features. At the word level, we examined lexical complexity and LLMs-preferred words to assess vocabulary usage and preference shifts. At the phrase level, we analyzed cohesion and readability to evaluate textual coherence and ease of comprehension. At the sentence level, we measured syntactic complexity and sentiment to capture structural variation and rhetorical tendencies in academic writing.

*3.2.1 Lexical Complexity*

(1) Lexical density

Lexical density is a crucial indicator of the balance between content words and function words within a text, with a higher proportion of content words generally indicating a more information-dense text. In this study, we employed the TAACS (Kyle & Crossley, 2015) to quantify both the types and tokens of content and function words. Additionally, we calculated the lexical density based on content word types and tokens.

(2) Lexical diversity

Lexical diversity is a measure of intricacy and variety of vocabulary in a text. The most widely used index in lexical diversity calculations is the Type-Token Ratio (TTR), which calculates the proportion of different lexical types to the total word tokens: $TTR = \frac{N_{types}}{N_{tokens}}$. To mitigate the influence of text length on calculations, we use moving-average TTR to calculate the lexical diversity of all abstracts, which was calculated by TAALED (Kyle et al., 2021).

(3) Lexical sophistication

Lexical sophistication refers to the depth and specialization of vocabulary，emphasizing the quality and complexity of words used in a text. In this study, we employed TAALES (Kyle & Crossley, 2015), a tool that utilizes data from the Corpus of Contemporary American English (COCA), which includes five registers: academic, fiction, magazine, news, and spoken. Given that this study aims to assess the impact of LLMs on academic writing, we selected the COCA-Academic register, as it is the most relevant to academic discourse, to measure both the range and frequency of vocabulary used.

*3.2.2 Syntactic complexity*

Syntactic complexity measures the complexity of sentence structure and syntactic features in language expression. Sentence length, clause length, and Minimal Terminable Unit (T-unit) counts are commonly used indicators of sentence complexity. In this study, we calculated four indices to measure the syntactic complexity of academic papers using TAASSC (Kyle, 2016), as shown in Table 3.

TABLE. 3. Syntactic complexity indices.

| **Index** | **Description** | **Formulas** |
|---|---|---|
| MLS | Mean length of sentence (average number of words per sentence) | $MLS = \frac{N_{word\ count}}{N_{sentence\ count}}$ |
| TU/S | T-units per sentence (ratio of T-units to sentences) | $TUS = \frac{N_{T-\ unit\ count}}{N_{sentence\ count}}$ |
| MLTU | Mean length of T-unit (average number of words per T-unit) | $MLTU = \frac{N_{word\ count}}{N_{T-\ unit\ count}}$ |
| MLC | Mean length of clause (average number of words per clause) | $MLC = \frac{N_{word\ count}}{N_{clasuse\ count}}$ |

***Note:** T-unit is the smallest syntactic unit that can stand alone as a complete thought. It consists of an independent clause along with all its dependent clauses.

*3.2.3 Readability*

Readability measures the ease with which readers can understand a paper's content. Papers with higher readability are more likely to be widely accepted and shared. In this study, we calculate the New Dale-Chall (NDC) and Flesch Reading Ease (FRE) indices to measure the readability of abstracts. The NDC formula assesses text difficulty based on a list of 3000 words recognized by fourth-grade American students, with words not included in this list are classified as difficult vocabulary. A higher NDC score indicates greater difficulty in reading. In contrast, the FRE calculates readability by analyzing sentence and word lengths, with a higher FRE score indicating easier readability. We used the textstat[4] package based on Python to calculate these two metrics with the following formulas:

$$NDC\ score\ =\ 0.1579\left(\frac{N\#difficult\ words}{N\#words}\times 100\right)+0.0496\left(\frac{N\#words}{N\#sentences}\right) \quad (1)$$

$$FRE\ score\ =\ 206.835-1.015\left(\frac{N\#words}{N\#sentences}\right)-84.6\left(\frac{N\#syllables}{N\#words}\right) \quad (2)$$

*3.2.4 Cohesion*

Cohesion reflects the logical flow and consistency between sentences in a text. In this study, we measure cohesion using three indices: lexical overlap, semantic overlap, and connectives, calculated with the TAACO tool (Crossley et al., 2016). Lexical overlap captures the repetition of words across sentences. Semantic overlap involves words and expressions with similar or related meanings in different sentences, and we use the Latent Semantic Analysis (LSA) to calculate the cosine similarity between adjacent sentences. Connectives reflect the proportion of different types of linking words used in the text. The specific indicators are shown in Table 4.

TABLE.4. Cohesion indices.

| **Index** | **Example** |
|---|---|
| Basic-connectives | for, and, nor |
| All logical | actually, admittedly, after all |
| All temporal | after, again |
| Reason and purpose | therefore, that is why, for this reason |
| Order | to begin with, next, first |

*3.2.5 Sentiment Analysis*

In academic writing, balancing emotional expression and objectivity is essential for helping readers understand the author's perspective. In this study, we use the Python-based natural language processing library Textblob[5] to measure the sentiment polarity and objectivity of arXiv abstracts. The library includes a sentiment lexicon, where each word has been assigned a polarity and subjectivity score, with normalization applied at the sentence level. This involved calculating polarity scores ranging from -1 to 1, with higher scores indicating a more positive tone. Furthermore, objectivity scores were calculated by subtracting the subjectivity scores from 1 (i.e., objectivity = 1 - subjectivity), resulting in a scale from 0 to 1, where higher values indicate the abstract maintains greater distance from personal perspectives.

*3.2.6 LLMs-preferred words*

With the deepening research on LLMs in academia, it's been observed that language generated by LLMs often follows standardized and normalized patterns. For example, words such as "innovative", "notable", and "excellent" are frequently used to describe research findings and methods. In this study, we use the word list developed by Liang et al. (2024a), which identifies the 100 most frequently used adjectives and adverbs by ChatGPT (gpt-4-0613). We analyze their usage by calculating the frequency in abstracts to explore LLMs' influence on academic writing. A detailed list of these words can be found in Table 5 of the appendix.

Table 6 summarizes the indices used in this study. Lexical complexity refers to the richness and variation of vocabulary in a text, and is measured in terms of lexical sophistication (use of advanced or uncommon words), lexical diversity (variety of different words), and lexical density (proportion of content words relative to function words). Syntactic complexity measures sentence structure complexity, including factors such as sentence length and the use of subordinate clauses. Cohesion refers to the grammatical and lexical relationships connecting elements within a text. Readability and sentiment measure the overall comprehensibility and authorial perspective of the abstract. LLMs-preferred words measure the relevance of the content to the use of the LLMs.

TABLE.6. Catalogue and Indicators for measuring academic writing style

| Category | Indicators | Variables |
|---|---|---|
| Lexical Complexity | / | Word count<br>Average word length |
| | Lexical density | content tokens, content types, function tokens, function types, density types, density tokens |
| | Lexical diversity | Mattr50 |
| | Lexical sophistication | Range_log_aw<br>Frequency_log_aw |
| Syntactic Complexity | / | Sentence count<br>Clause count<br>T-unit count |
| | Syntactic Complexity | MLC, MLTU, MLS, TU-S |
| Cohesion | Lexical overlap | Adjacent overlap all<br>Adjacent overlap argument |
| | Semantic overlap | LSA_all_sent |
| | Connectives | Basic-connectives |
| | | All logical |
| | | All temporal |
| | | Reason and purpose |
| | | Order |
| Readability | / | New Dale-Chall score<br>Flesch Reading Ease score |
| Sentiment | / | Polarity, Objectivity |
| LLMs-preferred words | / | Adverbs, Adjectives |

Among the measurement tools, TAALES, TAALED, TAACS, TAASSC, and TAACO were developed based on large-scale corpora and linguistic research, and their accuracy has been confirmed to show a high degree of consistency with human evaluations (Kyle, Crossley, & Berger, 2018; Crossley, Kyle, & McNamara, 2016), which helps ensure highly accurate and reliable results. To further validate the applicability of these tools to our dataset, we conducted a preliminary validation using 100 randomly selected abstracts. Each abstract was analyzed using implementations based on the Python toolkits NLTK and SpaCy, following the same vocabularies and computational methods described in the original documentation of each tool. The results showed that the average percentage errors were ±2.9% for TAALED, ±5.6% for TAASSC, ±7.1% for TAACO, ±5.3% for TAACS, and ±9.76% for TAALES. Considering potential differences in tokenization, sentence segmentation, part-of-speech tagging, and other implementation details when processing text, these margins of error still demonstrate strong agreement with the metadata and confirm the robustness of these tools in our context. More importantly, all those software can be easily accessed at https://www.linguisticanalysistools.org/ and can run on multiple operating systems, including Mac and Windows.

# 4 Results

*4.1 Overall changes in academic writing styles over time*

In this section, we analyzed the changes in the writing style of arXiv abstracts over the past decade and address RQ1. First, to explore the potential impact of ChatGPT on writing styles from a macro perspective, we analyzed changes in the number of new words and LLM-preferred words in abstracts from a random sample of 10,000 articles per time period.

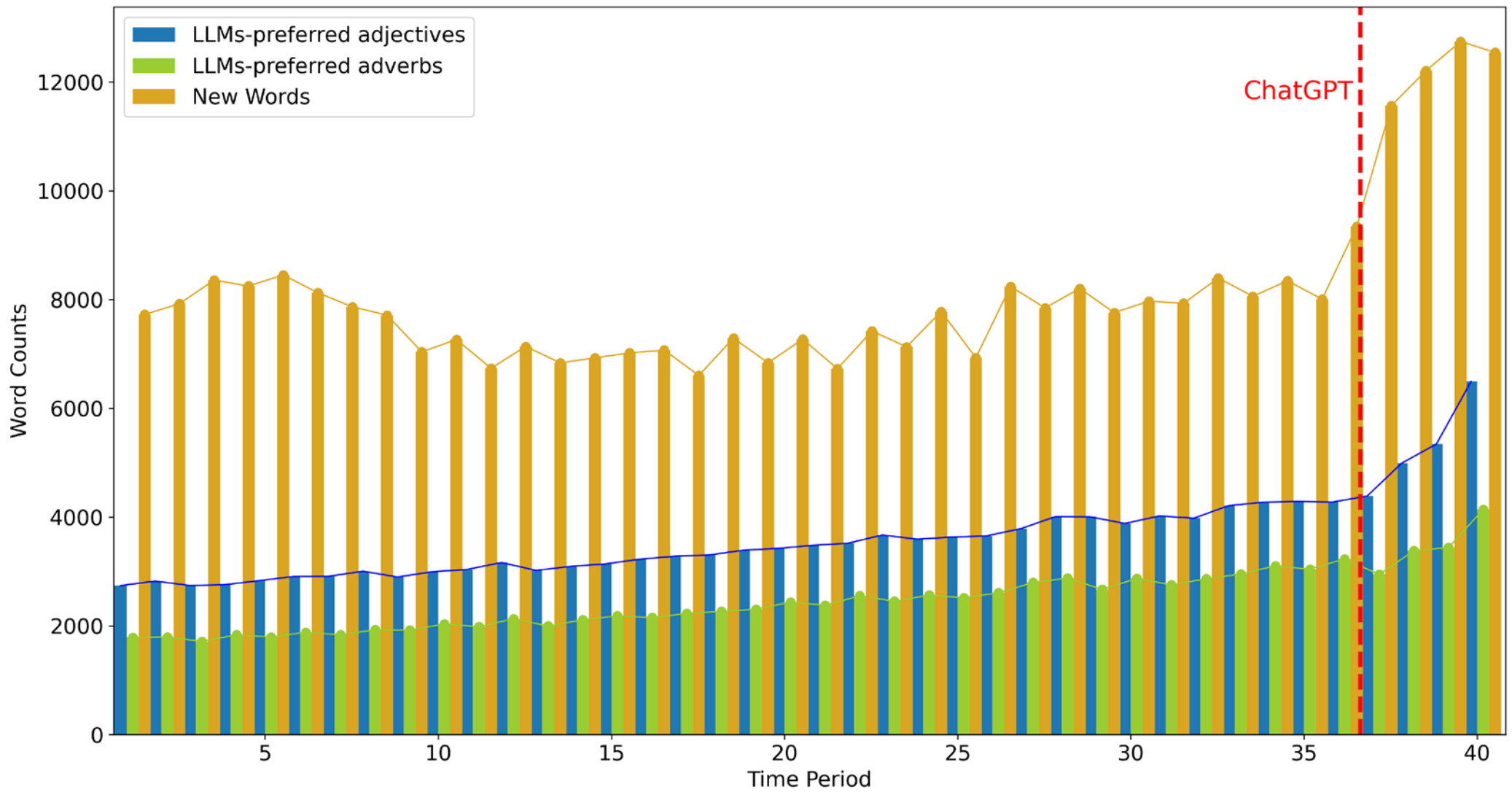

Fig. 2. New words and LLMs-preferred words trend in arXiv abstract over the past decade, with the x-axis representing the 40 time periods from 2014 to 2023.

As shown in Figure 2, there was a sharp increase in both following the release of ChatGPT. Considering that new words in scientific writing are often nouns, typically representing newly introduced concepts, technical terms, or methodologies, this surge suggests that the emergence of LLMs like ChatGPT has contributed to the introduction of novel technical vocabulary into academic abstracts. Moreover, the rise in LLM-preferred words, particularly the increased use of adverbs and adjectives, suggests that LLMs may not only shape the content of academic discourse but also its tone and expressiveness. For instance, as shown in Figure 3, adjectives like *"intricate"*, *"valuable"*, *"exceptional"*, *"pivotal"*, saw a sharp increase in usage following the release of ChatGPT, Similarly, adverbs like "*primarily*", "*thoroughly*", "*subsequently*", and "*particularly*", have become noticeably more frequent. The rise in LLMs-preferred words indicates a growing presence of language patterns commonly associated with such models (Geng & Trotta, 2024; Liang et al., 2024a).

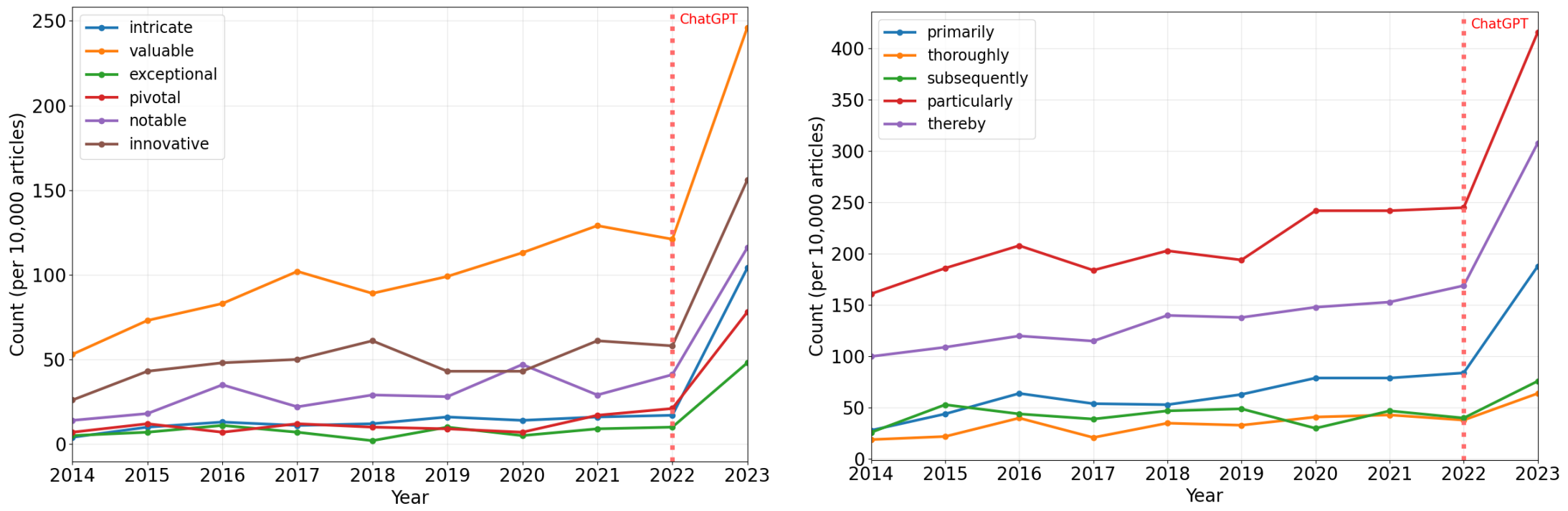

(a) adjectives (b) adverbs

FIG. 3. Adjectives and adverbs with significant increases in the field of Computer Science, with the x-axis representing the 40 time periods from 2014 to 2023.

Secondly, as illustrated in Figure 4, the past decade has witnessed an increase in both lexical density and lexical diversity (mattr50), while lexical sophistication has decreased. This trend suggests that although the vocabulary in abstracts has expanded and the range of topics has diversified, the overall words chosen tends to be simpler. Moreover, as abstracts have become longer, there has been a decline in syntactic complexity (MLC, MLS, TUS), indicating a trend toward clearer and more straightforward sentence structures. This shift aligns with the broader academic movement toward more transparent expression in in academic writing. However, following the release of ChatGPT, both average word length and lexical density have risen, indicating a more expansive and nuanced vocabulary in abstracts. Furthermore, the decline in syntactic complexity has accelerated, suggesting that the advent of LLMs has further streamlined sentence structures, which is crucial for the clear

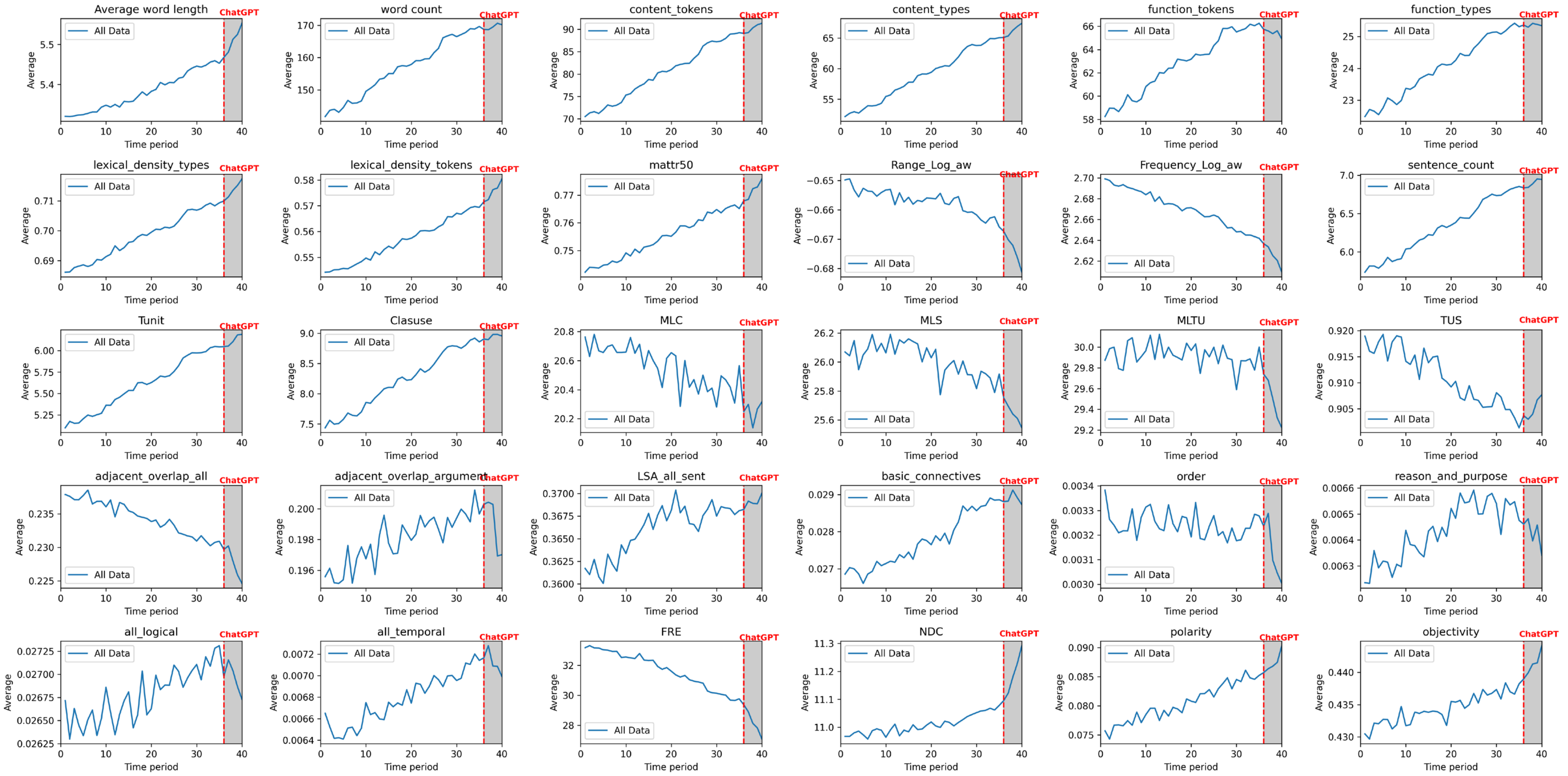

Fig. 4. Overall writing style changes of arXiv abstracts over the past decade, with the x-axis representing the 40 time periods from 2014 to 2023.

communication of knowledge.

Regarding cohesion, the results reveal a consistent increase in semantic overlap and the use of connectives over the past decade, indicating a decrease in repetitive sentences within abstracts and contributing to more coherent and logical expression of information. However, after the release of ChatGPT, there was a noticeable decline in both lexical overlap and basic connectives. This suggests that, while the content of abstracts has become more diverse and unique, the logical connections between sentences have weakened. One possible explanation is that LLMs often generate contextually appropriate connectives, which reduces the reliance on basic connectives that have traditionally helped to guide readers through academic texts.

Additionally, the readability of abstracts has declined over time, with this trend becoming more pronounced in 2023. This observation aligns with the finding of Plavén-Sigray et al. (2017) and Song et al. (2023). The readability score, calculated as the percentage of difficult words per sentence, likely reflects the increasing use of specialized terminology and concepts driven by advancements in science and various academic disciplines.

Finally, the results revealed a steady and gradual increase in both polarity and objectivity over time, with a marked acceleration in objectivity following the release of ChatGPT. This trend may reflect broader shifts in academic writing practices, as the emphasis on objectivity has grown in recent years. Given that many modern language models are designed to promote objective communication, their influence could be one factor contributing to the observed rise in objectivity scores over time.

Figure 5 shows the rate of change in writing style of arXiv abstracts from 2022 to 2023. The findings reveal a notable decline in readability (-5.6% in FRE score), suggesting that abstracts are becoming more challenging to comprehend. Additionally, the increase in NDC score (+1.2%) further supports this conclusion. The notable rise in positivity (+3.2%) reflects authors are adopting more optimistic language when emphasizing research findings and contributions. Additionally, macro-level indicators such as cohesion, sentiment and readability have shown more substantial changes compared to micro-level indicators like lexical complexity. To further evaluate the significance of the changes in writing style, we conducted a statistical analysis comparing the writing style in 2023 with that prior to the release of ChatGPT (2014–2022). For each writing style metric, we fitted a linear regression model to the 2014–2022 data to capture progressive trends and predicted the expected values for 2023. We then calculated the residuals, the differences between the actual values and the predicted trend, for both periods and performed a two-sample Kolmogorov–Smirnov (KS) test to compare their distributions. The KS test determines whether the 2023 residuals significantly deviate from the earlier years' distribution. If no significant change is observed, we expect relatively high *p-value*, indicating that the underlying distributions of the two groups are similar. The results, presented in Appendix Table 7, show that, 'LSA_all_sent' exhibited statistically significant changes with $p < 0.05$, while all the others showed even stronger significance with $p < 0.001$. These findings provide further support for the conclusion that the release of ChatGPT has contributed to a clear shift in writing style.

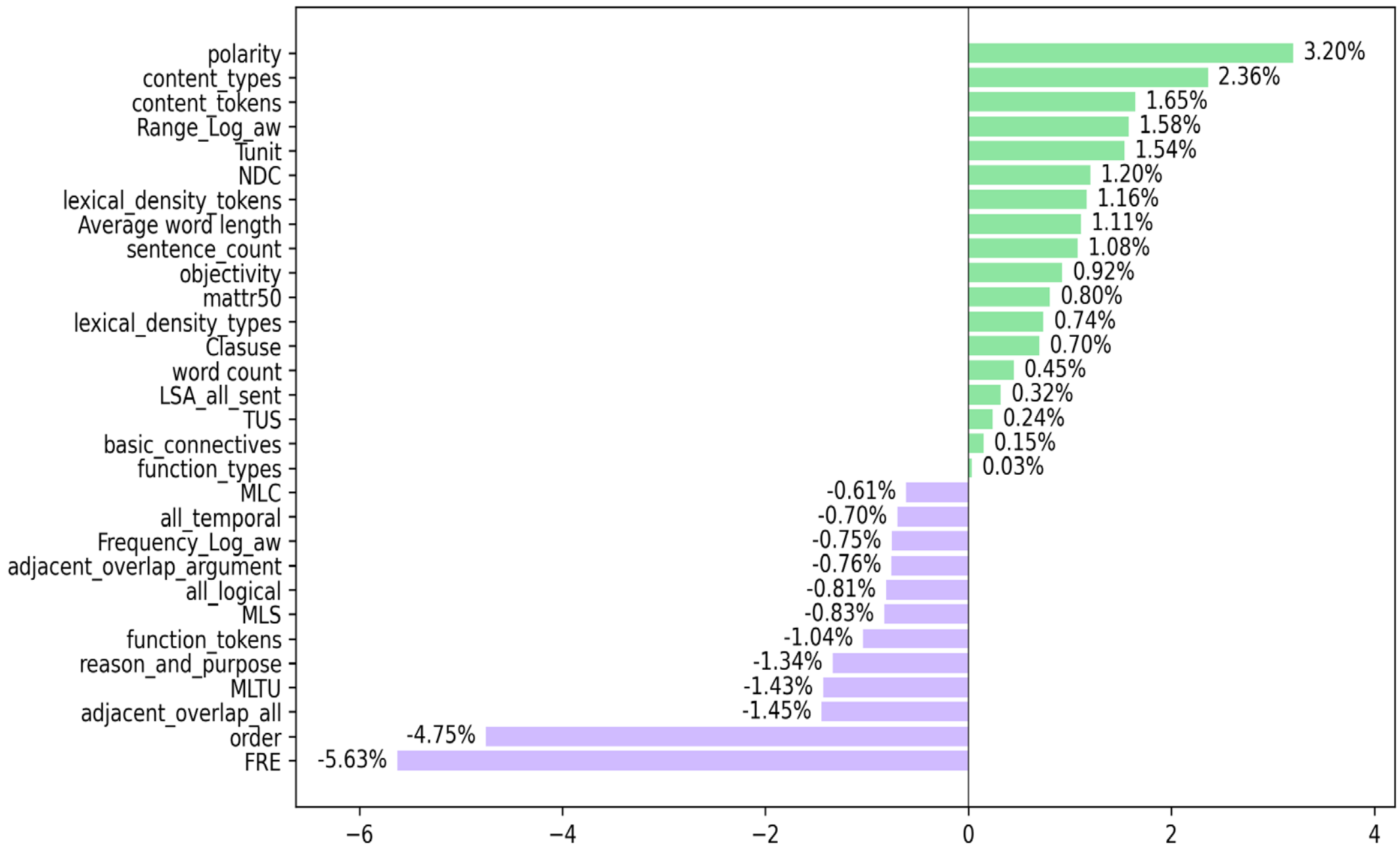

Fig.5. Writing style change rate in arXiv abstracts (2023 vs. 2022)

### *4.2 Writing styles variations of NESs and NNESs*

In this section, we analyzed the changes in writing styles among NESs and NNESs over the past decade, addressing RQ2. As shown in Figure 6, compared to NNESs, NESs demonstrate significantly higher in lexical complexity and cohesion (basic connectives, reason and purpose, logical, temporal) in arXiv abstracts. This suggesting that abstracts written by NESs tend to feature a richer and more flexible vocabulary, along with stronger sentence connectivity. However, since the introduction of ChatGPT, differences in lexical complexity, syntactic complexity, and cohesion have noticeably diminished. The convergence of metrics such as average word length, lexical density, T-unit, and mean length of clause (MLC) highlights the role of LLMs in narrowing the stylistic gap between NESs and NNESs, particularly enhancing the writing quality of NNESs to better align with the standards typically observed in NESs' work.

Furthermore, despite differences in lexical complexity and cohesion, the readability of abstracts between NESs and NNESs has remained largely similar, with both groups experiencing a decline over time. This decline may be linked to the essential role of the abstract in the article, which used to convey the main ideas and findings. Therefore, both NESs and NNESs will take the time to carefully to revise the abstract to ensure concise and logically presentations.

Lastly, the analysis reveals a trend of increasing polarity and objectivity in the writing of both NESs and NNESs over time. The release of ChatGPT appears to have accelerated the rise in objectivity scores, particularly among NNESs. This shift is likely due to NNESs facing more challenges in academic writing, which could drive a greater reliance on LLMs to improve language accuracy and fluency. Since most LLMs are designed to prioritize objectivity, their use likely contributed to the observed increase in objectivity scores.

To investigate the evolution of writing styles among scholars from diverse English-speaking backgrounds, we focused the six countries with the highest number of publications in 2023: China, India, Japan, France, the United States, and the Great Britain. Among these, the United States and the Great Britain are classified as NES countries, while China, Japan, and France are NNES countries. We analyzed the writing styles across four distinct time periods, both before and after the release of ChatGPT,

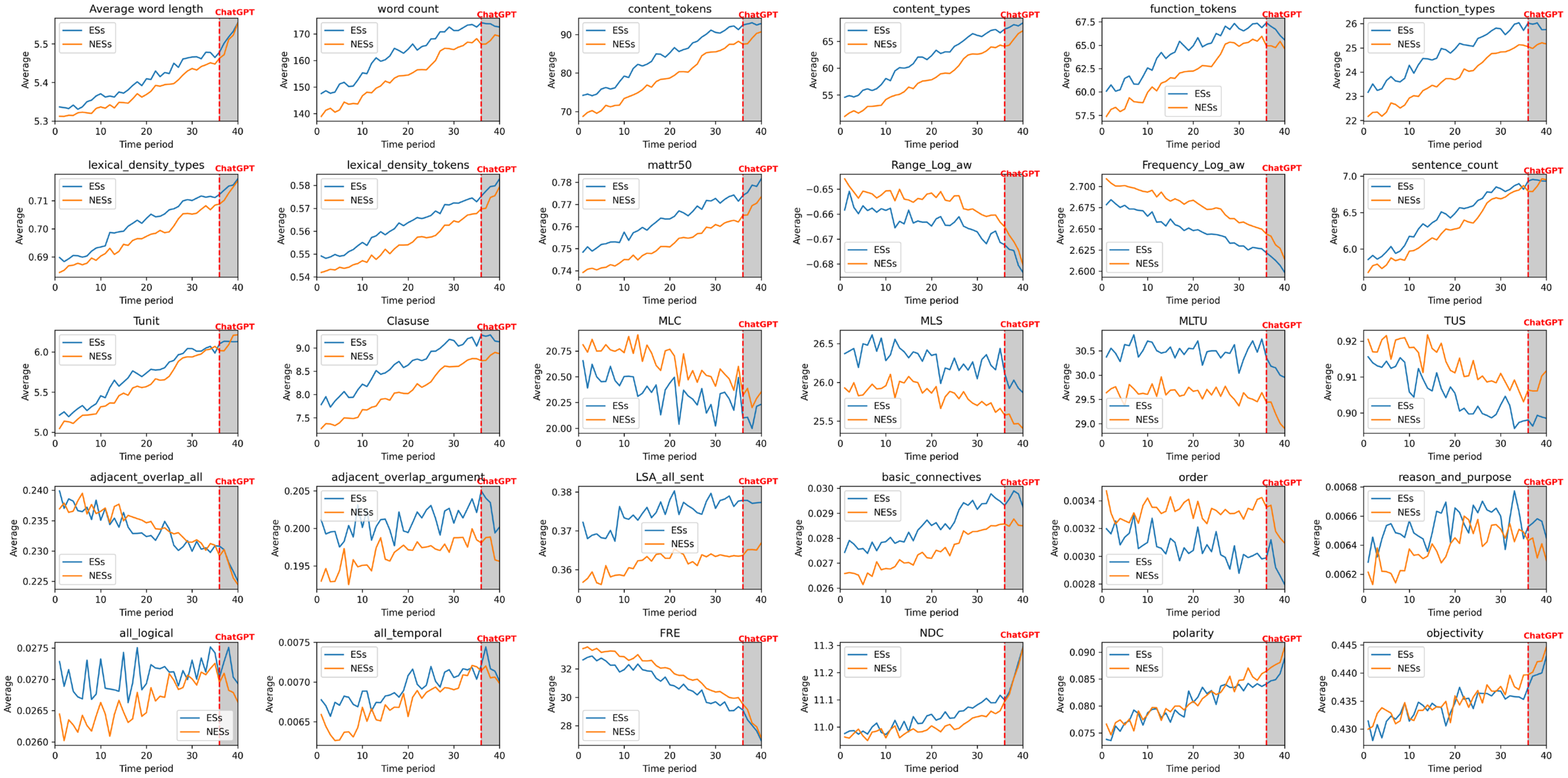

FIG. 6. Writing style change of NESs and NNESs in arXiv abstracts over the past decade, with the x-axis representing the 40 time periods from 2014 to 2023

as shown in Figure 7. The results show that compared to scholars from the United States and the Great Britain, authors from China, India, Japan, and France show more significant changes in lexical complexity, coherence, readability, and sentiment. Moreover, within the NNES countries, the changes observed in China, India, and Japan are more pronounced than those observed in France. This suggests a correlation between language environment on scholars in terms of shaping their writing style, with those from countries with weaker English proficiency showing more noticeable changes over time.

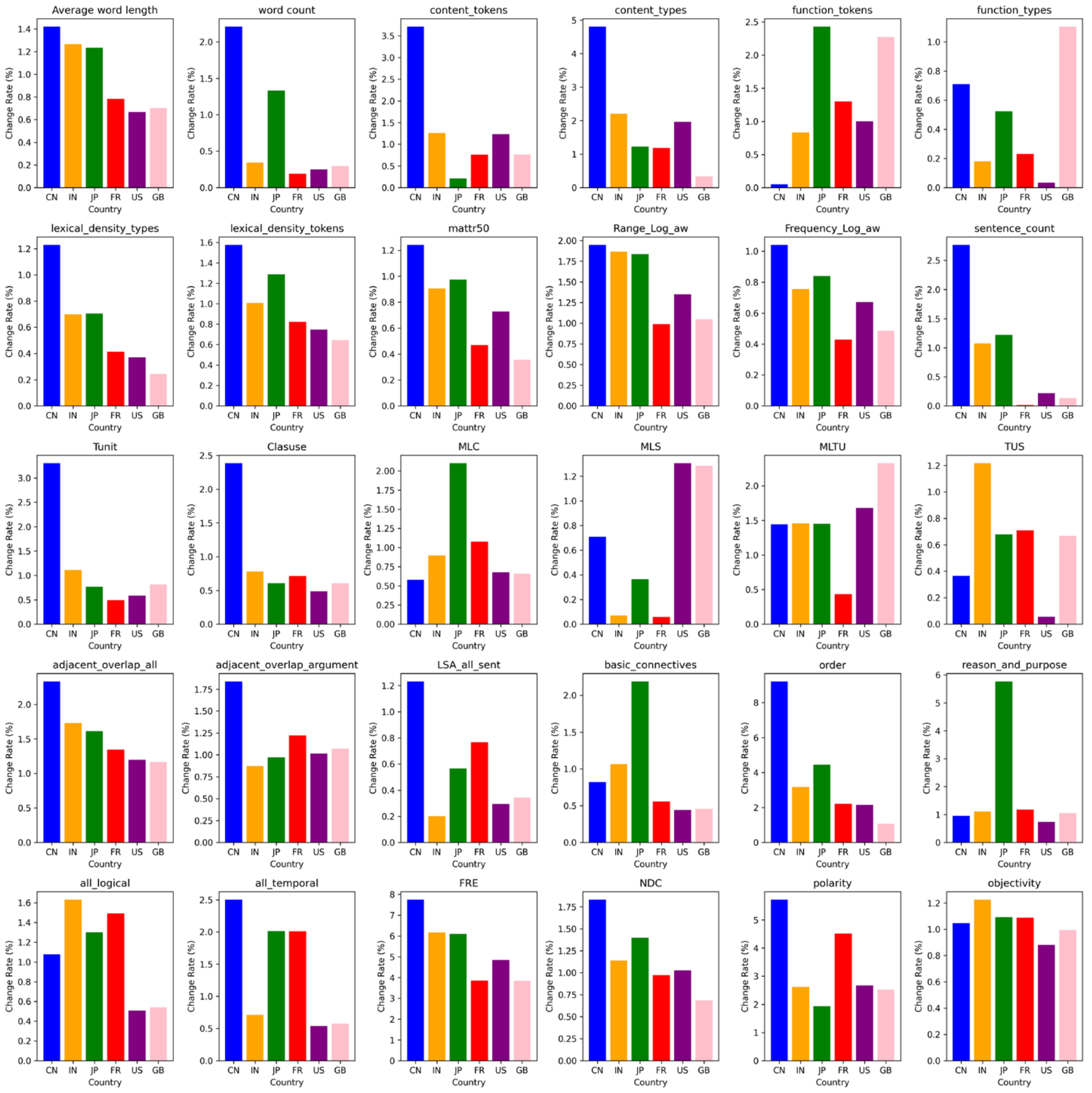

FIG.7. Differences in writing style changes among scholars from six countries in 2023 compared to 2022, where "CC"="China", "IN"="India", "JP"="Japan", "FR"="France", "US"="United States" and "GB"= "Great Britain".

Furthermore, we examined changes in the use of commonly used words and LLMs-preferred words in abstracts between NESs and NNESs. We focused on abstracts from the field of Computer Science for the years 2022 and 2023 for analysis because authors in these fields are more familiar to use these tools. Specifically, we randomly selected

TABLE. 8. Changes in commonly used words and LLMs-preferred words for NESs and NNESs after the release of ChatGPT.

| Word Type | Examples | NESs | | | NNESs | | |
|---|---|---|---|---|---|---|---|
| | | 2022 | 2023 | Δ | 2022 | 2023 | Δ |
| CC | and, or, but, either, nor | 58,270 | 59,851 | 2.71% | 56,339 | 58,318 | 3.51% |
| IN | of, in, for, with, on, by, from, at | 167,142 | 165,005 | -1.28% | 162,573 | 162,744 | -1.92% |
| VBZ | is, are, has, was | 39,579 | 37,819 | -4.45% | 39,943 | 36,652 | -8.24% |
| VBN | found, present, called, showed, proposed | 9,605 | 9,360 | -2.55% | 10,353 | 10,050 | -2.93% |
| RB | significantly, effectively, directly, automatically | 3,842 | 3,823 | -0.49% | 4,365 | 4,716 | 8.04% |
| NN | model, method, paper, work, approach, system | 2,7114 | 28798 | 6.21% | 28819 | 30974 | 7.18% |
| JJS | most, best, highest, largest, latest | 1,080 | 1,014 | -6.11% | 1,048 | 979 | -6.58% |
| LLMs-adjectives | comprehensive, innovative, notable, valuable | 3,567 | 4,058 | 12.09% | 3,675 | 4,169 | 13.44% |
| LLMs-adverbs | particularly, effectively, potentially, additionally | 3,788 | 4,070 | 7.44% | 3,886 | 4,242 | 9.16% |

***Note***: here ("CC", "Conjunctions"), ("IN", "Prepositions"), ("VBZ", "Present Tense Verb"), ("VBN", "Past Participle Verb"), ("RB", "Adverb"), ("NN", " Noun"), ("JJS", "Superlative Adjectives"), ("LLMs-adjectives", "High frequently adjectives used by LLMs") and ("LLMs- adverbs ", "High frequently adverbs used by LLMs). The numbers under NES and NNES represent the total frequency of the top 10 high-frequency words within each category.

10,000 abstracts per year from both NESs and NNESs and analyzed the parts of speech and frequencies of the words using NLTK library6. For the identification of common words, we first calculated the frequency of different words in the abstracts, then categorized them by part of speech, and selected the top ten high-frequency words from each category. These words were then combined for statistical analysis. The results, presented in Table 8, are evaluated with a 4% absolute change threshold, based on the observation that word category variations in the past five years (2017-2022) typically remained within this range. Based on the threshold, several word categories in Table 8 exhibited substantial changes between 2022 and 2023. In terms of increases, both NESs and NNESs showed notable rises in LLMs-adjectives (+12.09% for NESs; +13.44% for NNESs), LLMs-adverbs (+7.44% vs. +9.16%), and NN (+6.21% vs. +7.18%), indicating a broader adoption of descriptive expressions, particularly LLMs-adjectives such as "*comprehensive*", and "*innovative*", which are commonly used to modify task-related nouns like "*models*", "*methods*", and "*approach*". Additionally, RB (general adverbs) increased by +8.04% in NNESs, while remaining relatively stable in NESs (−0.49%). This larger shift among NNESs aligns with our earlier hypothesis that non-native speakers may have a greater reliance on LLMs to support their academic writing, leading to more pronounced stylistic changes in their abstracts. In terms of decreases, both groups exhibited declines in JJS (superlative adjectives), with 6.11% for NESs and 6.58% for NNESs, reflecting a move away from subjective emphasis through terms like "*most*" or "*best*". Similarly, the use of VBZ (present-tense verbs) decreased in both groups, by 4.45% in NESs and a sharper 8.24% in NNESs, indicating a tendency toward a more objective and passive style of writing.

*4.3 Writing styles variations across different disciplines*

In this section, we address the RQ3. Figure 8 shows the number of new words and LLMs-preferred words across disciplines over time. Notably, Computer Science exhibits a significant annual increase in new words, contrasting with the stable trends observed in Physics and Mathematics. This trend underscores the robust research activity and innovation within Computer Science. Over the past decade, pivotal models such as "*Transformer*", "*BERT*", and "*ChatGPT*" have gained widespread adoption in both academic and industrial settings. These advancements have established Computer Science as one of the most prominent research disciplines of the decade.

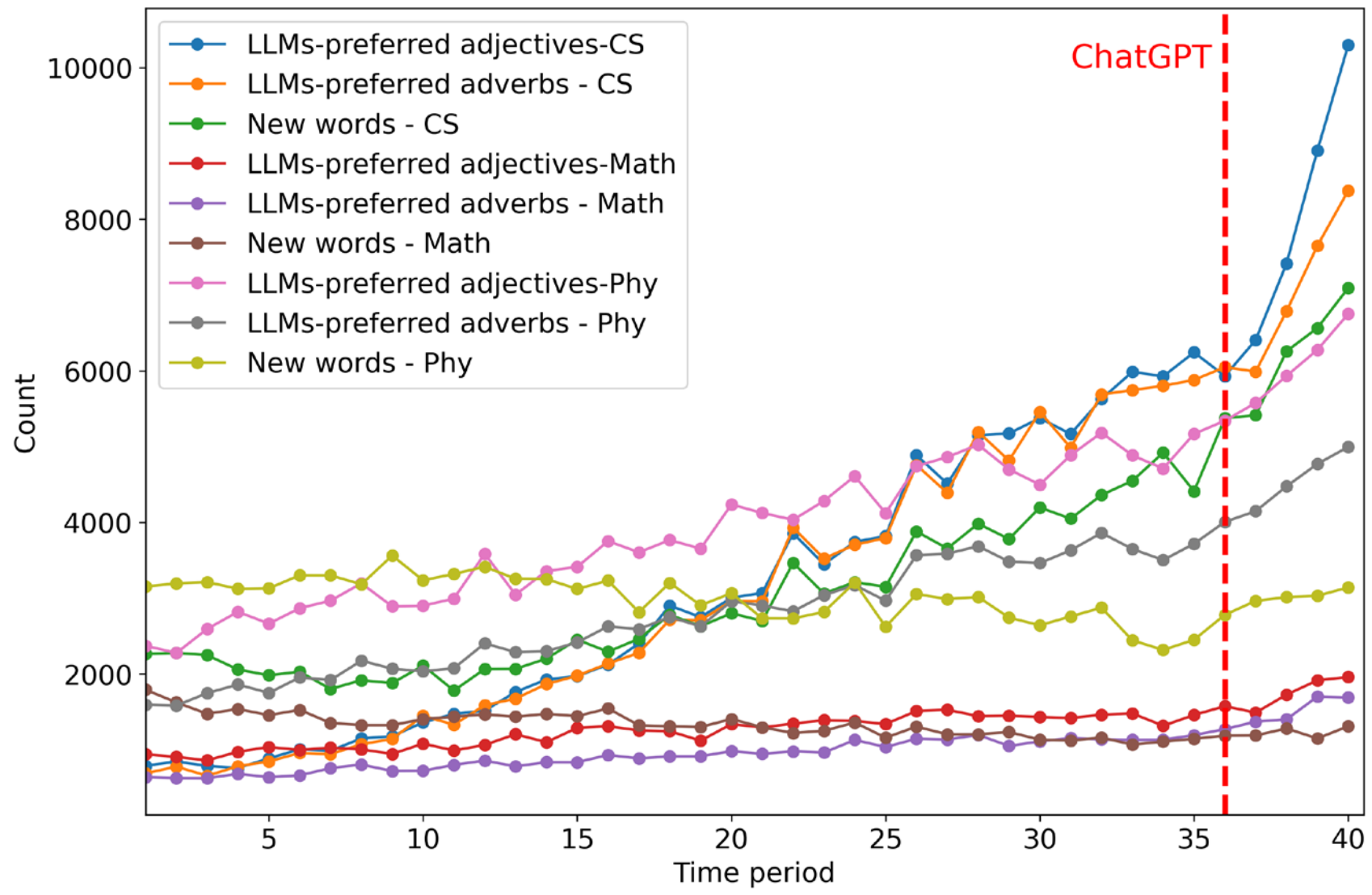

FIG.8. The number of new and LLMs-preferred words in different disciplines over time, with the x-axis representing the 40 time periods from 2014 to 2023.

Table 9 in the appendix lists the specific new words that have emerged in response to technological advancements across different disciplines. The results show that a total of 135,451 new words have emerged, with 76,823 in Computer Science, 44,531 in Physics, and 14,097 in Mathematics. In Computer Science, terms such as "*gpt-3.5-turbo*", "*controlnet*", and "*gpt-4v*" have appeared frequently following the rapid development of ChatGPT. Similarly, in Physics, new terms like "*LK-99*" and "*La3Ni2O7*" have appeared in relation to recent breakthroughs in superconductivity. In contrast, Mathematics has seen almost no such newly emerging terms. These findings suggest that the introduction of new words is closely tied to cutting-edge research and technological progress in each discipline.

As shown in Figure 9, following the release of ChatGPT, both Computer Science and Physics have exhibited an upward trend in LLMs-preferred words, whereas Mathematics has remained steady in this regard. One possible reason for this phenomenon is that, in this study, we selected LLMs-preferred words as adjectives and adverbs, both of which serve descriptive functions and are therefore relevant to descriptive tasks. To verify this point, we further examined the co-occurrence of LLMs-preferred words. As shown in the Figure 10, adjectives such as "*remarkable*", "*comprehensive*", "*effectively*", "*innovative*", and "*valuable*" are frequently used to emphasize the innovation of methods, their significance, and the improvement of results in Computer Science and Physics fields. In contrast, Mathematics, which primarily focuses on theoretical foundations and abstract concepts, relies less on such descriptive language. Meanwhile, Computer Science and Physics involve more descriptive tasks, such as explaining experimental results, illustrating models, and presenting detailed technical descriptions. These tasks often require the use of adjectives and adverbs to convey nuance and precision, which may explain the observed rise in LLMs-preferred words in these fields.

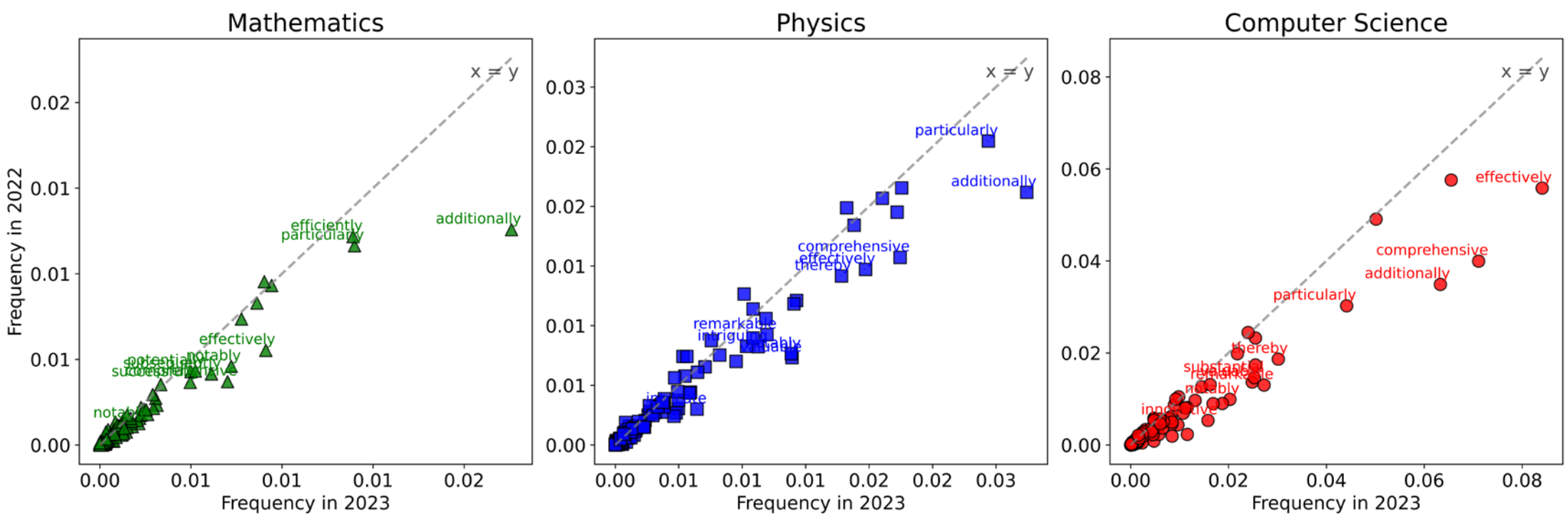

FIG.9. The frequency of LLM-preferred words in arXiv abstracts in 2023 and 2022, where the distribution of words below the diagonal line indicates an increase in frequency.

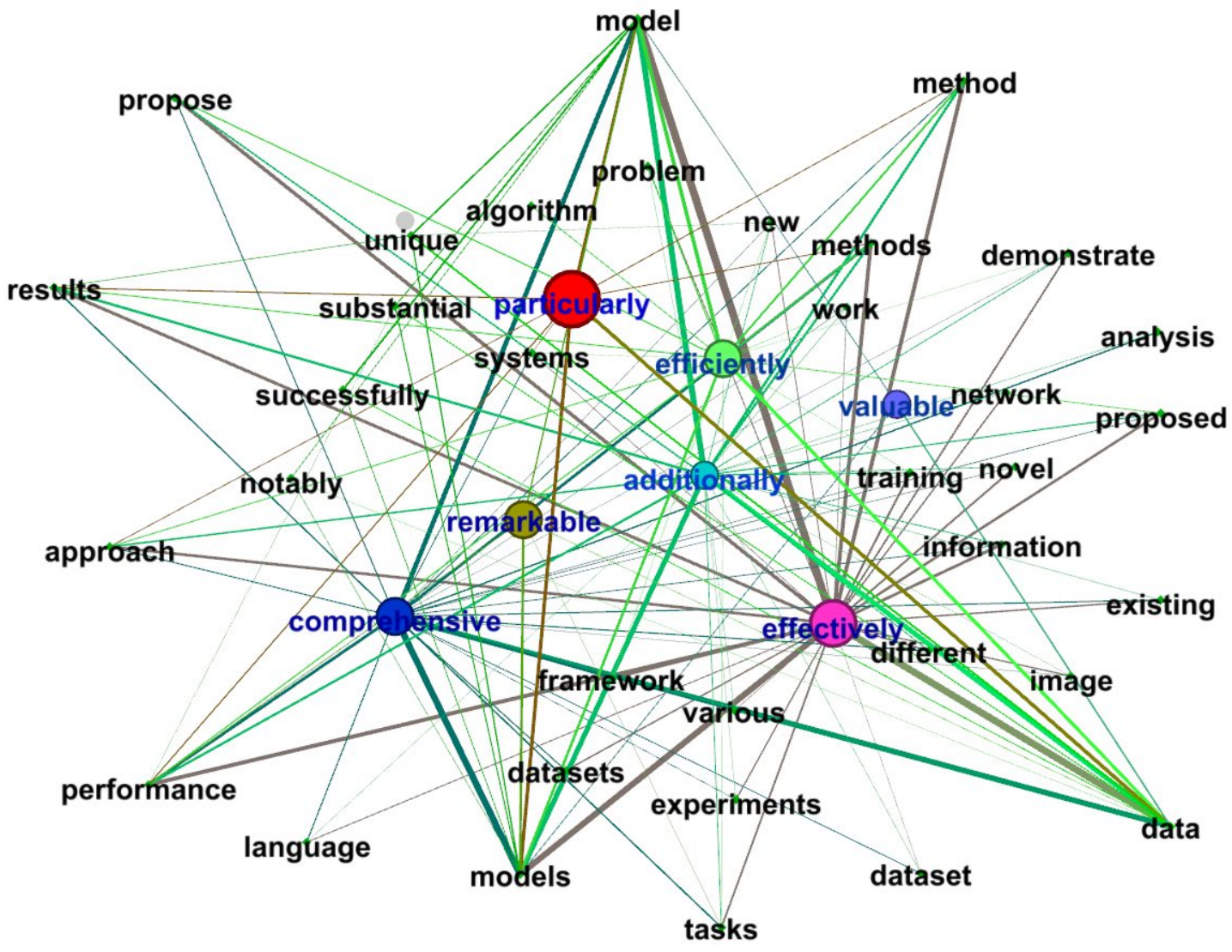

FIG.10. Example of co-occurrence words with LLMs-preferred words in arXiv abstracts from 2023, showing only words with a frequency greater than 1000.

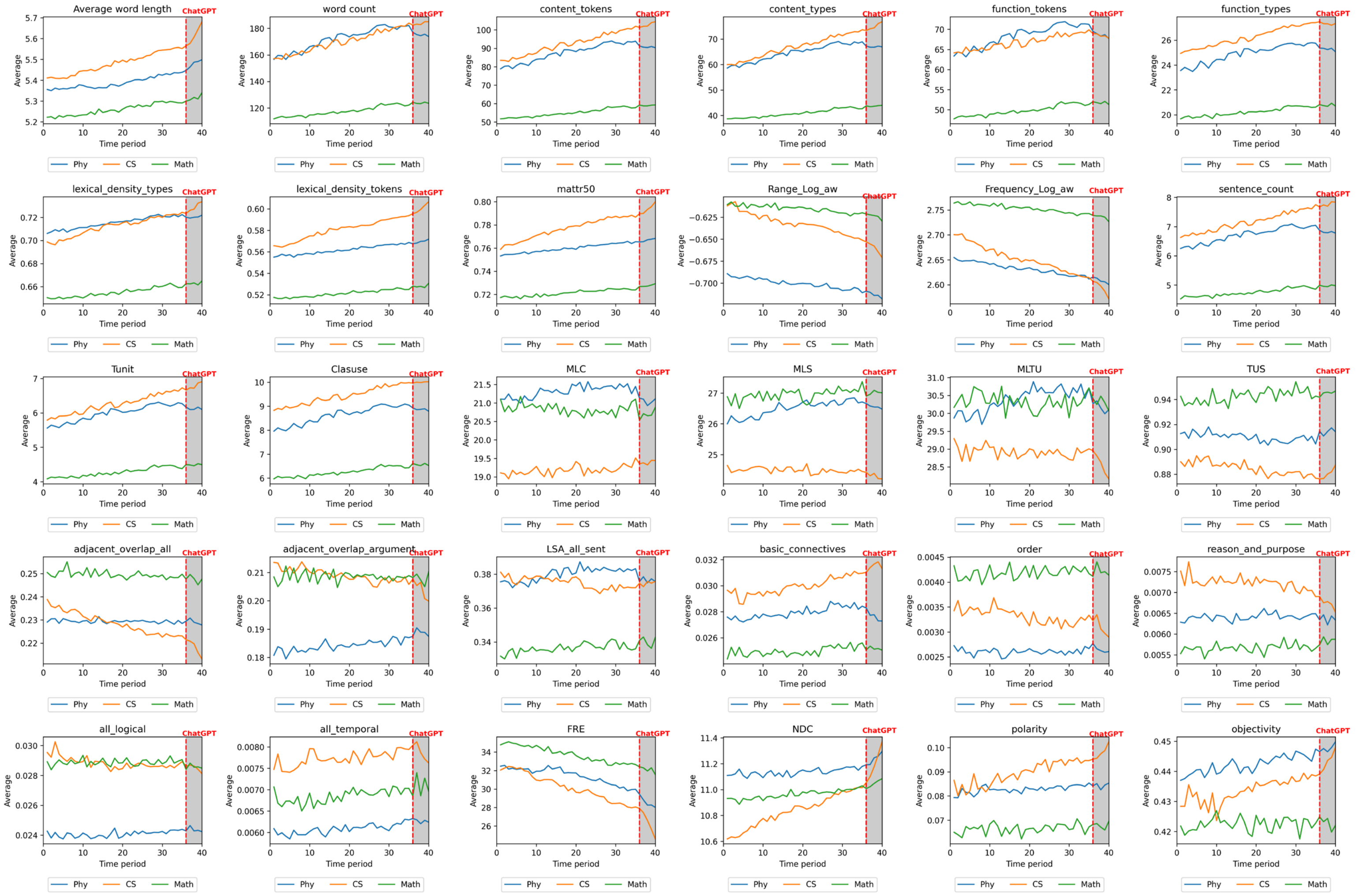

FIG. 11. Writing style changes across different disciplines in arXiv abstracts over the past decade, with the x-axis representing the 40 time periods from 2014 to 2023.

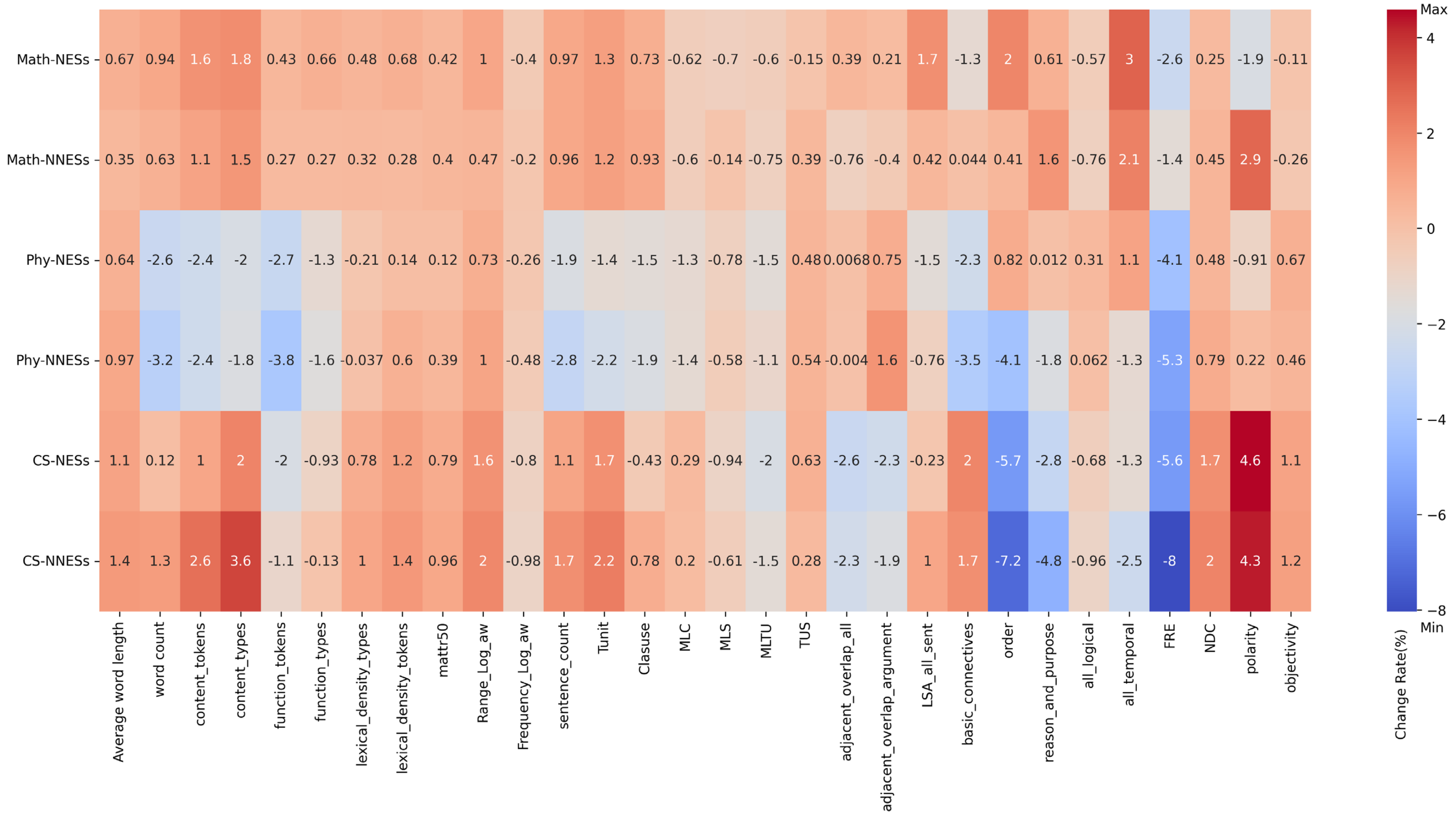

FIG.12. Change rate of writing styles across different disciplines in 2023 compared to 2022, with darker colors indicating greater rates of change.

Figure 11 illustrates the evolution of writing styles across disciplines over time. It is evident that Computer Science and Physics demonstrate higher lexical complexity and faster growth rates compared to Mathematics. Generally, the proportion of content words indicate the depth of knowledge within a discipline and the pace of innovation in the field. As interdisciplinary fields, Computer Science and Physics have witnessed a diverse array of vocabulary and concepts emerging in abstracts over time. In contrast, Mathematics, with its emphasis on theoretical derivation, shows relatively slower development. In terms of syntactic complexity and cohesion, all three disciplines have shown stability over time. However, following the introduction of ChatGPT, there was a decrease in adjacent overlaps and connectives in abstracts across all three disciplines.

Additionally, semantic overlap across disciplines has remained stable, suggesting that changes at the lexical level have not significantly impacted abstract semantics. Compared to Mathematics and Physics, Computer Science has experienced a more pronounced decline in syntactic complexity and cohesion. This trend may be due to Computer Science's closer integration with LLMs, enabling scholars in this field more effectively utilize these tools. Moreover, readability in Computer Science has decreased notably due to increased new words and average word length, while growth rates in Mathematics and Physics have remained stable. Regarding sentiment analysis, abstracts in Mathematics and Physics consistently exhibit polarity and objectivity. In contrast, abstracts in Computer Science have shown considerable improvement in polarity and objectivity following the release of ChatGPT.

Figure 12 presents a heatmap showing the rate of change in writing styles among NESs and NNESs across different disciplines from 2022 to 2023. Notably, an increasing trend is observed in lexical density in the field of Computer Science, particularly marked by a significant rise in the number of content words. This implies the emergence of new technical terms in this field after the release of ChatGPT. In addition, there is a downward trend in readability in Computer Science and Physics, which is closely related to the increase in lexical diversity and average word length. Furthermore, metrics in Mathematics have largely remained stable. Across language backgrounds, NNESs have shown more pronounced changes in writing style compared to NESs, with the most notable variations observed among NNESs in the field of Computer Science. However, within Mathematics, no significant changes in writing style were observed between NESs and NNESs, which supports the findings presented in section 4.2.

# 5 Discussion

*5.1Theoretical implications*

*5.1.1 Examining the Impact of LLMs on Academic Writing Style*

This paper analyzes the linguistic evolution of arXiv abstracts over the past decade. The results indicate a significant increase in adjectives and adverbs commonly associated with AI-generated text, a trend that appears to strongly correlate with the stylistic preferences observed in LLM outputs. This finding aligns with previous studies that evaluated the impact of LLMs from a probabilistic estimation perspective (Geng & Trotta, 2024; Liang et al., 2024a). From a linguistic perspective, we observe that lexical complexity, cohesion, and sensibility in academic writing have increased over time. Additionally, as abstract lengths have increased, sentence structures have simplified, reflected in the decline of syntactic complexity. However, following the release of ChatGPT, the introduction of new terms has further expanded lexical diversity, which has contributed to a notable decline in readability. Moreover, syntactic complexity decreased despite a decrease in the use of connectives, which may be attributed to LLMs generate contextually appropriate transitions rather than those widely used in academic writing. Finally, the observed rise in objectivity suggests a shift toward clearer and more neutral scientific communication. The above findings highlight the evolving influence of LLMs on academic writing and underscore the importance of balancing enhanced writing efficiency with the authenticity and domain-specific accuracy of scholarly content.

*5.1.2 Evolution changes of writing styles between NNESs and NESs*

This study investigated the changes in academic writing styles of NESs and NNESs over the past decade. We categorized first authors into two groups based on their country and analyzed their writing style changes following the introduction of ChatGPT. Prior to the release of ChatGPT, NESs demonstrated higher lexical complexity and cohesion but lower syntactic complexity, indicating greater proficiency in academic vocabulary usage and clarity of expression. However, these differences between NESs and NNESs have noticeably narrowed in the post-ChatGPT period. Furthermore, at the national level, scholars from NES countries such as China, India and Japan exhibited greater variation in their writing styles compared to those from the United States and the Great Britain. This variation could stem from differences in educational backgrounds, writing norms, or translation effects when scholars write in English as a second language. Despite concerns about potential academic integrity issues associated with the use of LLM-based writing tools (Else, 2023), the observed changes suggest that LLMs can significantly improve the writing skills of NNESs, helping to reduce language barriers in academic publishing and fostering greater equality and collaboration in scientific research.

*5.1.3 The changes of writing styles in different disciplines*

The differences in writing styles across disciplines may be attributed to variations in research objects, methods and the acceptance of new technologies and tools. Through an analysis of the differences in writing styles between Computer Science, Physics, and Mathematics, we found that Computer Science, as a technology-driven field, is more inclined to adopt and experiment with new tools like ChatGPT, leading to greater stylistic variation. In contrast, Mathematics, as a fundamental discipline, has maintained a consistently stable writing style. This indicates that disciplinary characteristics significantly shape the evolution of writing styles. This observation aligns with the theory of technology diffusion, which posits that new technologies spread through distinct adoption stages (Siler & Larivière, 2024; Zhou et al., 2023). In academic writing, researchers from different disciplines may embrace LLMs at different stages, resulting in disparate trends of change in their writing styles.

*5.2 Practical implications*

Through a large-scale and multiple-perspectives analysis, this study reveals significant shifts in academic writing styles in the era of LLMs, suggesting their widespread adoption in academic writing. Notably, for NNESs, LLMs offer a promising solution to enhance writing fluency within a relatively short period. Moreover, the impact of LLMs varies across disciplines, with hard sciences like Computer Science, Physics, and Mathematics exhibiting distinct stylistic changes, providing a linguistic perspective on the influence of emerging technologies across fields. Given these changes, journals in disciplines heavily influenced by LLMs should consider updating their writing standards and academic norms. At the same time, researchers must balance efficiency gains with maintaining the novelty and authenticity of their work to avoid potential academic integrity issues. Finally, it is important to note that the excessive reliance on these tools may lead to academic misconduct, such as plagiarism and fabrication. Through comparative analysis, existing studies (Goulart et al., 2024; Sardinha, 2024) have highlighted differences between AI-generated and human-authored academic texts in terms of information production, narrative structure, and expression. The future integration of these linguistic features with the linguistic changes observed in our study, along with LLM-preferred words and LLM-detection tools, may enhance the ability to screen scholarly submissions for plagiarism.

### *5.3 Limitations*

This study explores the impact of ChatGPT on the academic writing style and provides valuable insights, but we acknowledge that there are some limitations. First, the analysis focuses on preprint abstracts, while the abstract is an important part of the paper, it cannot fully reflect the overall characteristics of academic writing. Second, abstracts were categorized based on the native language of the first author, but the linguistic backgrounds of co-authors could also influence writing style. Moreover, using the institutional affiliation of the first author to infer their language background is not entirely accurate, as authors may publish from institutions in countries different from their native ones. Finally, although existing studies have demonstrated the widespread effects of LLMs on academic writing, the causal relationship between LLM use and changes in academic writing requires further validation through inference and additional parameter testing.

## 6 Conclusion and Future Works

This study explores the changes in academic writing styles following the release of ChatGPT. By analyzing 823,798 arXiv abstracts across more than 30 linguistic indicators, our results demonstrate significant impacts of LLMs on academic writing. Specifically, following the advent of ChatGPT, a significant increase in lexical complexity and a decline in readability and sentence complexity were observed across the three academic disciplines. This suggests that LLMs introduced more new terminology terms, making abstracts more difficult to understand while simplified sentence structure. Moreover, cohesion and the frequency of common words decreased, whereas the use of LLM-preferred words rose sharply. Furthermore, scholars with weaker English proficiency are more likely to use LLMs for academic writing, and the changes in the macro-level indicators (cohesion, readability, sentiment) are more obvious than those in micro-level (lexical complexity) indicators, which suggests that scholars pay more attention to the using LLMs to improve the overall fluency and logic of their abstracts. Finally, Computer Science shows a higher presence of new words and LLMs-preferred words, with more noticeable changes in writing style compared to Physics and Mathematics. This suggests that Computer Science scholars may be more inclined to adopt and integrate LLMs into their writing practices, while the impact on Mathematics appears to be relatively limited.

In future work, we plan to download PDF versions of the papers and parse out the full-text to comprehensively measure changes in academic writing style. Furthermore, we will consider the contributions and linguistic background of all authors as grouping factors and examine how writing style changes for the same authors over time, aiming to enhance the robustness of our analyses.

**Acknowledgements**

This paper is supported by the National Natural Science Foundation of China (Grant No. 72074113), the Science Fund for Creative Research Group of the National Natural Science Foundation of China (No. 71921002). We also appreciate the anonymous reviewers and the editor for their valuable comments, which have helped improve the quality of this paper.

**Declarations**

Conflict of interest: The author(s) declared no potential conflicts of interest with respect to the research, authorship, and/or publication of this article.

## REFERENCES

AlAfnan, M. A., & MohdZuki, S. F. (2023). Do Artificial Intelligence Chatbots Have a Writing Style? An Investigation into the Stylistic Features of ChatGPT-4. *Journal of Artificial Intelligence and Technology*, *3*(3), Article 3. https://doi.org/10.37965/jait.2023.0267

Altmäe, S., Sola-Leyva, A., & Salumets, A. (2023). Artificial intelligence in scientific writing: A friend or a foe? *Reproductive BioMedicine Online*, *47*(1), 3–9. https://doi.org/10.1016/j.rbmo.2023.04.009

Atkinson, D. (1998). *Scientific Discourse in Sociohistorical Context: The Philosophical Transactions of the Royal Society of London, 1675-1975*. Routledge. https://doi.org/10.4324/9781410601704

Bennett, K. (2009). English academic style manuals: A survey. *Journal of English for Academic Purposes*, *8*(1), 43–54. https://doi.org/10.1016/j.jeap.2008.12.003

Biber, D., & Gray, B. (2010). Challenging stereotypes about academic writing: Complexity, elaboration, explicitness. *Journal of English for Academic Purposes*, *9*(1), 2–20. https://doi.org/10.1016/j.jeap.2010.01.001

Cao, Y., Li, S., Liu, Y., Yan, Z., Dai, Y., Yu, P. S., & Sun, L. (2023). *A Comprehensive Survey of AI-Generated Content (AIGC): A History of Generative AI from GAN to ChatGPT* (arXiv:2303.04226). arXiv. https://doi.org/10.48550/arXiv.2303.04226

Chen, B., Deng, D., Zhong, Z., & Zhang, C. (2020). Exploring linguistic characteristics of highly browsed and downloaded academic articles. *Scientometrics*, *122*(3), 1769–1790. https://doi.org/10.1007/s11192-020-03361-4

Corrêa Jr, E. A., Silva, F. N., Costa, L. D. F., & Amancio, D. R. (2017). Patterns of authors contribution in scientific manuscripts. *Journal of Informetrics*, *11*(2), 498-510.

Covington, M. A., & McFall, J. D. (2010). Cutting the Gordian knot: The moving-average type–token ratio (MATTR). *Journal of quantitative linguistics*, *17*(2), 94-100.

Crossley, S. A., Kyle, K., & McNamara, D. S. (2016). The tool for the automatic analysis of text cohesion (TAACO): Automatic assessment of local, global, and text cohesion. *Behavior Research Methods*, *48*(4), 1227–1237. https://doi.org/10.3758/s13428-015-0651-7

Dong, S., Mao, J., Ke, Q., & Pei, L. (2024). Decoding the writing styles of disciplines: A large-scale quantitative analysis. *Information Processing & Management*, *61*(4), 103718. https://doi.org/10.1016/j.ipm.2024.103718

Else, H. (2023). Abstracts written by ChatGPT fool scientists. *Nature*, *613*(7944), 423–423. https://doi.org/10.1038/d41586-023-00056-7

Flowerdew, J. (1999). Problems in writing for scholarly publication in English: The case of Hong Kong. *Journal of Second Language Writing*, *8*(3), 243–264. https://doi.org/10.1016/S1060-

3743(99)80116-7

Gao, C. A., Howard, F. M., Markov, N. S., Dyer, E. C., Ramesh, S., Luo, Y., & Pearson, A. T. (2023). Comparing scientific abstracts generated by ChatGPT to real abstracts with detectors and blinded human reviewers. *Npj Digital Medicine*, *6*(1), 75. https://doi.org/10.1038/s41746-023-00819-6

Geng, M., & Trotta, R. (2024). *Is ChatGPT Transforming Academics' Writing Style?* (arXiv:2404.08627). arXiv. http://arxiv.org/abs/2404.08627

Gidiotis, A., & Tsoumakas, G. (2020). *A Divide-and-Conquer Approach to the Summarization of Long Documents* (arXiv:2004.06190). arXiv. http://arxiv.org/abs/2004.06190

Goulart, L., Matte, M. L., Mendoza, A., Alvarado, L., & Veloso, I. (2024). AI or student writing? Analyzing the situational and linguistic characteristics of undergraduate student writing and AI-generated assignments. *Journal of Second Language Writing*, 66, 101160.

Gök, A., & Karaulova, M. (2024). How "international" is international research collaboration? *Journal of the Association for Information Science and Technology*, *75*(2), 97–114. https://doi.org/10.1002/asi.24842

Graham, F. (2023). Daily briefing: ChatGPT listed as author on research papers. *Nature*. https://doi.org/10.1038/d41586-023-00188-w

Greenhalgh, T., Robert, G., Macfarlane, F., Bate, P., & Kyriakidou, O. (2004). Diffusion of innovations in service organizations: systematic review and recommendations. *The milbank quarterly*, 82(4), 581-629.

Gross, A. G., Harmon, J. E., & Reidy, M. S. (2002). *Communicating Science: The Scientific Article from the 17th Century to the Present*. Oxford University Press.

Gui, Q., Liu, C., & Du, D. (2019). Globalization of science and international scientific collaboration: A network perspective. *Geoforum*, *105*, 1–12. https://doi.org/10.1016/j.geoforum.2019.06.017

Hwang, T., Aggarwal, N., Khan, P. Z., Roberts, T., Mahmood, A., Griffiths, M. M., ... & Khan, S. (2024). Can ChatGPT assist authors with abstract writing in medical journals? Evaluating the quality of scientific abstracts generated by ChatGPT and original abstracts. *Plos one*, 19(2), e0297701.

Hu, H., Wang, D., & Deng, S. (2021). Analysis of the scientific literature's abstract writing style and citations. *Online Information Review*, *45*(7), 1290–1305. https://doi.org/10.1108/OIR-05-2020-0188

Huang, J. C. (2010). Publishing and learning writing for publication in English: Perspectives of NNES PhD students in science. *Journal of English for Academic Purposes*, *9*(1), 33–44. https://doi.org/10.1016/j.jeap.2009.10.001

Huang, J., & Tan, M. (2023). The role of ChatGPT in scientific communication: Writing better scientific review articles. *American Journal of Cancer Research*, *13*(4), 1148–1154.

Juzek, T. S., & Ward, Z. B. (2025). Why Does ChatGPT "Delve" So Much? Exploring the Sources of Lexical Overrepresentation in Large Language Models. In *Proceedings of the 31st International Conference on Computational Linguistics* (pp. 6397-6411).

Kyle, K. (2016). Measuring Syntactic Development in L2 Writing: Fine Grained Indices of Syntactic Complexity and Usage-Based Indices of Syntactic Sophistication. *Applied Linguistics and English as a Second Language Dissertations*. https://doi.org/10.57709/8501051

Kyle, K., & Crossley, S. A. (2015). Automatically Assessing Lexical Sophistication: Indices, Tools, Findings, and Application. *TESOL Quarterly*, *49*(4), 757–786. https://doi.org/10.1002/tesq.194

Kyle, K., Crossley, S., & Berger, C. (2018). The tool for the automatic analysis of lexical sophistication (TAALES): Version 2.0. *Behavior research methods*, 50, 1030-1046.

Kyle, K., & Crossley, S. A. (2018). Measuring Syntactic Complexity in L2 Writing Using Fine-Grained Clausal and Phrasal Indices. *The Modern Language Journal*, *102*(2), 333–349. https://doi.org/10.1111/modl.12468

Kyle, K., Crossley, S. A., & Jarvis, S. (2021). Assessing the Validity of Lexical Diversity Indices Using Direct Judgements. *Language Assessment Quarterly*, *18*(2), 154–170. https://doi.org/10.1080/15434303.2020.1844205

Larivière, V., Desrochers, N., Macaluso, B., Mongeon, P., Paul-Hus, A., & Sugimoto, C. R. (2016). Contributorship and division of labor in knowledge production. *Social Studies of Science*, *46*(3), 417–435. https://doi.org/10.1177/0306312716650046

Liang, W., Izzo, Z., Zhang, Y., Lepp, H., Cao, H., Zhao, X., Chen, L., Ye, H., Liu, S., Huang, Z., McFarland, D. A., & Zou, J. Y. (2024a). *Monitoring AI-Modified Content at Scale: A Case Study on the Impact of ChatGPT on AI Conference Peer Reviews* (arXiv:2403.07183). arXiv. http://arxiv.org/abs/2403.07183

Liang, W., Zhang, Y., Wu, Z., Lepp, H., Ji, W., Zhao, X., Cao, H., Liu, S., He, S., Huang, Z., Yang, D., Potts, C., Manning, C. D., & Zou, J. Y. (2024b). *Mapping the Increasing Use of LLMs in Scientific Papers* (arXiv:2404.01268). arXiv. http://arxiv.org/abs/2404.01268

Liu, J., & Bu, Y. (2024). Towards the relationship between AIGC in manuscript writing and author profiles: evidence from preprints in LLMs. arXiv preprint arXiv:2404.15799.

Lozić, E., & Štular, B. (2023). ChatGPT v Bard v Bing v Claude 2 v Aria v human-expert. How good are AI chatbots at scientific writing? *Future Internet*, *15*(10), 336. https://doi.org/10.3390/fi15100336

Lu, C., Bu, Y., Wang, J., Ding, Y., Torvik, V., Schnaars, M., & Zhang, C. (2019). Examining scientific writing styles from the perspective of linguistic complexity. *Journal of the Association for Information Science and Technology*, *70*(5), 462–475. https://doi.org/10.1002/asi.24126

Lu, C., Zhang, C., Xiao, C., & Ding, Y. (2022). Contributorship in scientific collaborations: The perspective of contribution-based byline orders. *Information Processing & Management*, *59*(3), 102944.

Lu, X. (2010). Automatic analysis of syntactic complexity in second language writing. *International Journal of Corpus Linguistics*, *15*(4), 474–496. https://doi.org/10.1075/ijcl.15.4.02lu

Ma, Y., Liu, J., Yi, F., Cheng, Q., Huang, Y., Lu, W., & Liu, X. (n.d.). *AI vs. Human—Differentiation Analysis of Scientific Content Generation*.

Melitz, J., & Toubal, F. (2014). Native language, spoken language, translation and trade. *Journal of International Economics*, *93*(2), 351–363. https://doi.org/10.1016/j.jinteco.2014.04.004

Mundt, K., & Groves, M. (2016). A double-edged sword: The merits and the policy implications of Google Translate in higher education. *European Journal of Higher Education*, *6*(4), 387–401. https://doi.org/10.1080/21568235.2016.1172248

Nur Fitria, T. (2021). "Grammarly" as AI-powered English Writing Assistant: Students' Alternative for English Writing. *Metathesis Journal of English Language Literature and Teaching*, *5*,

65–78. https://doi.org/10.31002/metathesis.v5i1.3519

Onal, S., & Kulavuz-Onal, D. (2024). A Cross-Disciplinary Examination of the Instructional Uses of ChatGPT in Higher Education. *Journal of Educational Technology Systems*, *52*(3), 301–324. https://doi.org/10.1177/00472395231196532

Ortega, L. (2003). Syntactic Complexity Measures and their Relationship to L2 Proficiency: A Research Synthesis of College-level L2 Writing. *Applied Linguistics*, *24*(4), 492–518. https://doi.org/10.1093/applin/24.4.492

Plavén-Sigray, P., Matheson, G. J., Schiffler, B. C., & Thompson, W. H. (2017). The readability of scientific texts is decreasing over time. *eLife*, *6*, e27725. https://doi.org/10.7554/eLife.27725

Rogers, E. M., Singhal, A., & Quinlan, M. M. (2014). Diffusion of innovations. In An integrated approach to communication theory and research (pp. 432-448). Routledge.

Ribeiro, L. C., Rapini, M. S., Silva, L. A., & Albuquerque, E. M. (2018). Growth patterns of the network of international collaboration in science. *Scientometrics*, *114*(1), 159–179. https://doi.org/10.1007/s11192-017-2573-x

Sardinha, T. B. (2024). AI-generated vs human-authored texts: A multidimensional comparison. *Applied Corpus Linguistics*, 4(1), 100083.

Siler, K., & Larivière, V. (2024). Varieties of diffusion in academic publishing: How status and legitimacy influence growth trajectories of new innovations. *Journal of the Association for Information Science and Technology*, *75*(2), 132–151. https://doi.org/10.1002/asi.24844

Snow, C. E. (2010). Academic Language and the Challenge of Reading for Learning About Science. *Science*, *328*(5977), 450–452. https://doi.org/10.1126/science.1182597

Song, N., Chen, K., & Zhao, Y. (2023). Understanding writing styles of scientific papers in the IS-LS domain: Evidence from abstracts over the past three decades. *Journal of Informetrics*, *17*(1), 101377. https://doi.org/10.1016/j.joi.2023.101377

Thorp, H. H. (2023). ChatGPT is fun, but not an author. *Science*, *379*(6630), 313–313. https://doi.org/10.1126/science.adg7879

Uzun, L. (2023). ChatGPT and Academic Integrity Concerns: Detecting Artificial Intelligence Generated Content. *Language Education and Technology*, *3*(1), Article 1. http://www.langedutech.com/letjournal/index.php/let/article/view/49

Wang, G., Wang, H., Sun, X., Wang, N., & Wang, L. (2023). Linguistic complexity in scientific writing: A large-scale diachronic study from 1821 to 1920. *Scientometrics*, *128*(1), 441–460. https://doi.org/10.1007/s11192-022-04550-z

Zhou, H., Guns, R., & Engels, T. C. E. (2023). Towards indicating interdisciplinarity: Characterizing interdisciplinary knowledge flow. *Journal of the Association for Information Science and Technology*, *74*(11), 1325–1340. https://doi.org/10.1002/asi.24829

## Appendix:

**TABLE 5. Top100 frequently used adjectives and adverbs by ChatGPT.**

| Top100 adjectives disproportionately used more frequently by ChatGPT | | | | |
|---|---|---|---|---|
| commendable | innovative | meticulous | intricate | notable |
| versatile | noteworthy | invaluable | pivotal | potent |
| fresh | ingenious | cogent | ongoing | tangible |
| profound | methodical | laudable | lucid | appreciable |
| fascinating | adaptable | admirable | refreshing | proficient |
| intriguing | thoughtful | credible | exceptional | digestible |
| prevalent | interpretative | remarkable | seamless | economical |
| proactive | interdisciplinary | sustainable | optimizable | comprehensive |
| vital | pragmatic | comprehensible | unique | fuller |
| authentic | foundational | distinctive | pertinent | valuable |
| invasive | speedy | inherent | considerable | holistic |
| insightful | operational | substantial | compelling | technological |
| beneficial | excellent | keen | cultural | unauthorized |
| strategic | expansive | prospective | vivid | consequential |
| manageable | unprecedented | inclusive | asymmetrical | cohesive |
| replicable | quicker | defensive | wider | imaginative |
| traditional | competent | contentious | widespread | environmental |
| instrumental | substantive | creative | academic | sizeable |
| versatile | noteworthy | invaluable | pivotal | potent |
| fresh | ingenious | cogent | ongoing | tangible |

| Top100 adverbs disproportionately used more frequently by ChatGPT | | | | |
|---|---|---|---|---|
| meticulously | reportedly | lucidly | innovatively | aptly |
| methodically | excellently | compellingly | impressively | undoubtedly |
| scholarly | strategically | intriguingly | competently | intelligently |
| hitherto | thoughtfully | profoundly | undeniably | admirably |
| creatively | logically | markedly | thereby | contextually |
| distinctly | judiciously | cleverly | invariably | successfully |
| chiefly | refreshingly | constructively | inadvertently | effectively |
| intellectually | rightly | convincingly | comprehensively | seamlessly |
| predominantly | coherently | evidently | notably | professionally |
| subtly | synergistically | productively | purportedly | remarkably |
| traditionally | starkly | promptly | richly | nonetheless |
| elegantly | smartly | solidly | inadequately | effortlessly |
| forth | firmly | autonomously | duly | critically |
| immensely | beautifully | maliciously | finely | succinctly |
| further | robustly | decidedly | conclusively | diversely |
| exceptionally | concurrently | appreciably | methodologically | universally |
| thoroughly | soundly | particularly | elaborately | uniquely |
| neatly | definitively | substantively | usefully | adversely |
| primarily | principally | discriminatively | efficiently | scientifically |
| alike | herein | additionally | subsequently | potentially |

**TABLE 7. Statistical analysis of writing style changes in arXiv abstracts in 2023.**

| Metric | $R^2$ | 2023 Prediction Mean | 2023 True Mean | K-S Statistic | Cohen's d | *P-value* |
|---|---|---|---|---|---|---|
| Average word length | 0.969 | 5.476 | 5.520 | 0.046 | 0.307 | $p < 0.001$*** |
| word count | 0.977 | 173.424 | 169.789 | 0.039 | 0.182 | $p < 0.001$*** |
| content_tokens | 0.984 | 91.835 | 90.524 | 0.031 | 0.261 | $p < 0.001$*** |
| content_types | 0.984 | 66.959 | 66.547 | 0.029 | 0.292 | $p < 0.001$*** |
| function_tokens | 0.964 | 67.511 | 65.357 | 0.045 | 0.081 | $p < 0.001$*** |
| function_types | 0.983 | 25.815 | 25.360 | 0.056 | 0.160 | $p < 0.001$*** |
| lexical_density_types | 0.974 | 0.713 | 0.714 | 0.032 | 0.245 | $p < 0.001$*** |
| lexical_density_tokens | 0.979 | 0.573 | 0.577 | 0.039 | 0.322 | $p < 0.001$*** |
| mattr50 | 0.978 | 0.770 | 0.773 | 0.023 | 0.275 | $p < 0.001$*** |
| Range_Log_aw | 0.735 | -0.665 | -0.675 | 0.045 | -0.153 | $p < 0.001$*** |
| Frequency_Log_aw | 0.964 | 2.636 | 2.622 | 0.042 | -0.260 | $p < 0.001$*** |
| sentence_count | 0.982 | 7.003 | 6.910 | 0.080 | 0.192 | $p < 0.001$*** |
| Tunit | 0.981 | 6.186 | 6.135 | 0.085 | 0.190 | $p < 0.001$*** |
| Clasuse | 0.981 | 9.124 | 8.954 | 0.061 | 0.168 | $p < 0.001$*** |
| MLC | 0.648 | 20.318 | 20.254 | 0.031 | -0.039 | $p < 0.001$*** |
| MLS | 0.513 | 25.830 | 25.620 | 0.030 | -0.049 | $p < 0.001$*** |
| MLTU | 0.096 | 29.836 | 29.420 | 0.025 | -0.046 | $p < 0.001$*** |
| TUS | 0.863 | 0.902 | 0.905 | 0.325 | -0.019 | $p < 0.001$*** |
| adjacent_overlap_all | 0.925 | 0.230 | 0.227 | 0.026 | -0.094 | $p < 0.001$*** |
| adjacent_overlap_argument | 0.763 | 0.201 | 0.199 | 0.031 | 0.002 | $p < 0.001$*** |
| LSA_all_sent | 0.702 | 0.370 | 0.369 | 0.020 | 0.022 | $p < 0.05$* |
| basic_connectives | 0.934 | 0.029 | 0.029 | 0.023 | 0.068 | $p < 0.001$*** |
| order | 0.074 | 0.003 | 0.003 | 0.661 | -0.025 | $p < 0.001$*** |
| reason_and_purpose | 0.693 | 0.007 | 0.006 | 0.397 | -0.005 | $p < 0.001$*** |
| all_logical | 0.717 | 0.027 | 0.027 | 0.029 | 0.007 | $p < 0.001$*** |
| all_temporal | 0.904 | 0.007 | 0.007 | 0.334 | 0.036 | $p < 0.001$*** |
| FRE | 0.966 | 29.252 | 27.947 | 0.046 | -0.246 | $p < 0.001$*** |
| NDC | 0.790 | 11.075 | 11.209 | 0.065 | 0.179 | $p < 0.001$*** |
| polarity | 0.896 | 0.087 | 0.088 | 0.015 | 0.067 | $p < 0.001$*** |
| objectivity | 0.831 | 0.435 | 0.439 | 0.018 | 0.060 | $p < 0.001$*** |

Note: *$p < 0.05$, ***$p < 0.001$.

**TABLE 9. Top 100 new words emerging in arXiv abstracts (10,000 articles per discipline) in 2023.**

| Words | Discipline | #N | Words | Discipline | #N |
|---|---|---|---|---|---|
| gpt-3.5-turbo | Computer Science | 65 | vall-e | Computer Science | 8 |
| controlnet | Computer Science | 61 | audioldm | Computer Science | 8 |
| gpt-4v | Computer Science | 59 | 3327 | Physics | 8 |
| llama2 | Computer Science | 37 | llms-generated | Computer Science | 8 |
| llava | Computer Science | 35 | vicuna-13b | Computer Science | 8 |
| lk-99 | Physics | 35 | q-former | Computer Science | 8 |
| yolov8 | Computer Science | 33 | stable-diffusion | Computer Science | 8 |
| gpt3.5 | Computer Science | 32 | tree-of-thought | Computer Science | 8 |
| jailbreak | Computer Science | 32 | 33b | Computer Science | 8 |
| chatgpt-generated | Computer Science | 30 | llama-2-chat | Computer Science | 8 |
| llama-7b | Computer Science | 29 | llama2-7b | Computer Science | 8 |
| promptable | Computer Science | 27 | llama-2-7b | Computer Science | 8 |
| lvlms | Computer Science | 27 | fmmm | Physics | 7 |
| blip-2 | Computer Science | 26 | graviton-photon | Physics | 7 |
| chatgpt-based | Computer Science | 21 | satech-01 | Physics | 7 |
| dinov2 | Computer Science | 20 | avsbench | Computer Science | 7 |
| llm-driven | Computer Science | 19 | laion-2b | Computer Science | 7 |
| lvlm | Computer Science | 19 | sa-1b | Computer Science | 7 |
| chatgpt-3.5 | Computer Science | 18 | human-llm | Computer Science | 7 |
| jailbreaking | Computer Science | 18 | jjem | Physics | 7 |
| starcoder | Computer Science | 17 | wizardlm | Computer Science | 7 |
| chatgpt-4 | Computer Science | 16 | imagebind | Computer Science | 7 |
| palm-2 | Computer Science | 15 | llm-empowered | Computer Science | 7 |
| 10-x | Physics | 15 | sboms | Computer Science | 6 |
| semeval-2023 | Computer Science | 14 | sbom | Computer Science | 6 |
| minigpt-4 | Computer Science | 14 | prompt-enhanced | Computer Science | 6 |
| jailbreaks | Computer Science | 14 | close-sourced | Computer Science | 6 |
| self-instruct | Computer Science | 13 | zs-cir | Computer Science | 6 |
| llm-guided | Computer Science | 13 | nuimages | Computer Science | 6 |
| la3ni2o7 | Physics | 13 | mode-pairing | Physics | 6 |
| llms-based | Computer Science | 12 | llama-65b | Computer Science | 6 |
| qlora | Computer Science | 12 | gecam-c | Physics | 6 |
| hex-p | Physics | 12 | nonstabilizerness | Physics | 6 |
| dreamfusion | Computer Science | 11 | democratise | Computer Science | 6 |
| babylm | Computer Science | 11 | zero-1-to-3 | Computer Science | 6 |
| llama-13b | Computer Science | 11 | flan-t5-xxl | Computer Science | 6 |
| chat-gpt | Computer Science | 11 | gptzero | Computer Science | 6 |
| instructpix2pix | Computer Science | 11 | cvpr2023 | Computer Science | 6 |
| instruction-guided | Computer Science | 11 | evol-instruct | Computer Science | 6 |
| intelligence-generated | Computer Science | 11 | kit-ml | Computer Science | 6 |
| 230307a | Physics | 10 | point-e | Computer Science | 6 |
| text-3d | Computer Science | 10 | 127-qubit | Physics | 6 |
| chatgpt-like | Computer Science | 10 | ml-superb | Computer Science | 6 |
| segment-anything | Computer Science | 10 | partimagenet | Computer Science | 6 |
| llama-based | Computer Science | 10 | gpt-4-based | Computer Science | 6 |
| sdxl | Computer Science | 10 | ml-ready | Computer Science | 6 |
| tool-augmented | Computer Science | 9 | ovod | Computer Science | 6 |
| instruction-response | Computer Science | 9 | mt-bench | Computer Science | 6 |
| vicuna-7b | Computer Science | 9 | qgem | Physics | 6 |

**Endnotes**

[1] https://arxiv.org/
[2] https://www.kaggle.com/Cornell-University/arxiv
[3] https://docs.openalex.org/
[4] https://pypi.org/project/textstat/
[5] https://pypi.org/project/textblob/
[6] https://www.nltk.org/